\documentclass[a4paper,fleqn]{cas-dc}

\usepackage[authoryear]{natbib}
\usepackage{booktabs}
\usepackage{tabularx}
\usepackage{array}
\usepackage{xcolor}
\usepackage{microtype}

\usepackage{amsmath,amssymb}
\usepackage{graphicx}
\usepackage{rotating}
\usepackage{textcomp}
\usepackage{lineno}
\usepackage{etoolbox}
\hypersetup{hidelinks}

\def\tsc#1{\csdef{#1}{\textsc{\lowercase{#1}}\xspace}}
\tsc{WGM}
\tsc{QE}

\begin{document}
\let\WriteBookmarks\relax
\def\floatpagefraction{1}
\def\textpagefraction{.001}

\renewcommand{\figurename}{Fig.}
\newcolumntype{L}{>{\raggedright\arraybackslash}X}
\newcolumntype{Y}{>{\centering\arraybackslash}X}
\newcommand{\PICP}{\mathrm{PICP}}

\shorttitle{Uncertainty quantification under graded distribution shift}

\shortauthors{A. Lawal et al.}

\title [mode = title]{Benchmarking uncertainty quantification for engineering regression under graded distribution shift}



%

\author[1]{Ahmad Lawal}


\cormark[1]


\credit{Conceptualization, Methodology, Software, Formal analysis, Writing -- original draft, Writing -- review \& editing}

\affiliation[1]{organization={School of Computer Science and Informatics, De Montfort University},
            city={Leicester},
            country={UK}}

\author[1]{Rahim Pasbanigoloojeh}
\credit{Validation, Writing -- review \& editing}

\author[1]{Asmau Lawal Bilbis}
\credit{Writing -- review \& editing}







\cortext[1]{Corresponding author}



\begin{abstract}
Machine-learning models in engineering are often used beyond their training conditions, where reliable uncertainty quantification (UQ) matters most. Existing comparisons of regression UQ methods rely on random splits, which assess calibration only in distribution. We introduce a three-rung shift ladder (in distribution, soft shift and hard shift) built along physically meaningful shift variables, and evaluate 12 methods from five families on seven public engineering datasets: five extrapolation settings, an unseen turbofan fault mode and a winter air-quality sensor deployment. All tuning uses source data only. In distribution, conformal methods cover at least 0.946 of test points on every dataset. Under hard shift, mean coverage falls to 0.40--0.75 across the 11 methods that always return an interval, and no systematic difference in method ranks is detected (Friedman $p = 0.63$). Weighted conformalised quantile regression stays near nominal on the points it answers, but abstains on 85\% of hard-shift points on average and covers only 0.67 of the few Concrete points it answers. Source-validation gains predict improvement in distribution and under soft shift, but not under hard shift (Spearman $\rho = 0.04$). Engineering evaluations should split calibration data by deployment unit, report subgroup coverage and sharpness, and treat abstention as an explicit operational outcome.
\end{abstract}




\maketitle











\section{Introduction}\label{sec:intro}

Engineering decisions are routinely made on predictions outside the conditions a model was trained on. A mix design is approved on a long-term strength predicted from early-age tests. A power plant schedules output for a heatwave hotter than any day in its operating record. A maintenance plan relies on remaining-useful-life estimates for an engine developing a fault the model has never seen. A low-cost air-quality sensor, calibrated against a reference instrument over the summer, is left to report through the winter. In each case the prediction is most needed exactly where the training data are thinnest or absent.

Uncertainty quantification (UQ) is meant to be the safeguard. A model that reports a prediction interval rather than a single number lets the engineer see when it does not know, and a well-calibrated 95\% interval should contain the true value 95\% of the time. The methods for producing such intervals are now mature and widely used in reliability and surrogate modelling. They include Gaussian processes (GPs), Bayesian and ensemble neural networks, distributional and quantile regression, and conformal prediction, which wraps any model and guarantees marginal coverage in finite samples when calibration and test points are exchangeable. The question this paper asks is whether these intervals keep their promise where they are needed, beyond the training envelope.

The evidence so far mostly cannot answer it. Comparative studies of UQ for regression, including the closest to this work \citep{dewolf2023valid}, evaluate methods on random train--test splits of standard datasets. A random split makes training and test points exchangeable by construction. That is precisely the condition under which conformal methods guarantee coverage and under which other methods are easiest to calibrate. A random-split benchmark therefore measures in-distribution calibration, but it does not reveal whether uncertainty estimates remain reliable as deployment moves away from the training distribution. Studies that do introduce distribution shift have largely done so for classification, with synthetic corruptions \citep{ovadia2019can}, rather than for engineering regression with physically meaningful shifts.

\begin{figure}[pos=tbp]
\centering
\includegraphics[width=\linewidth,trim=0 17 0 19,clip]{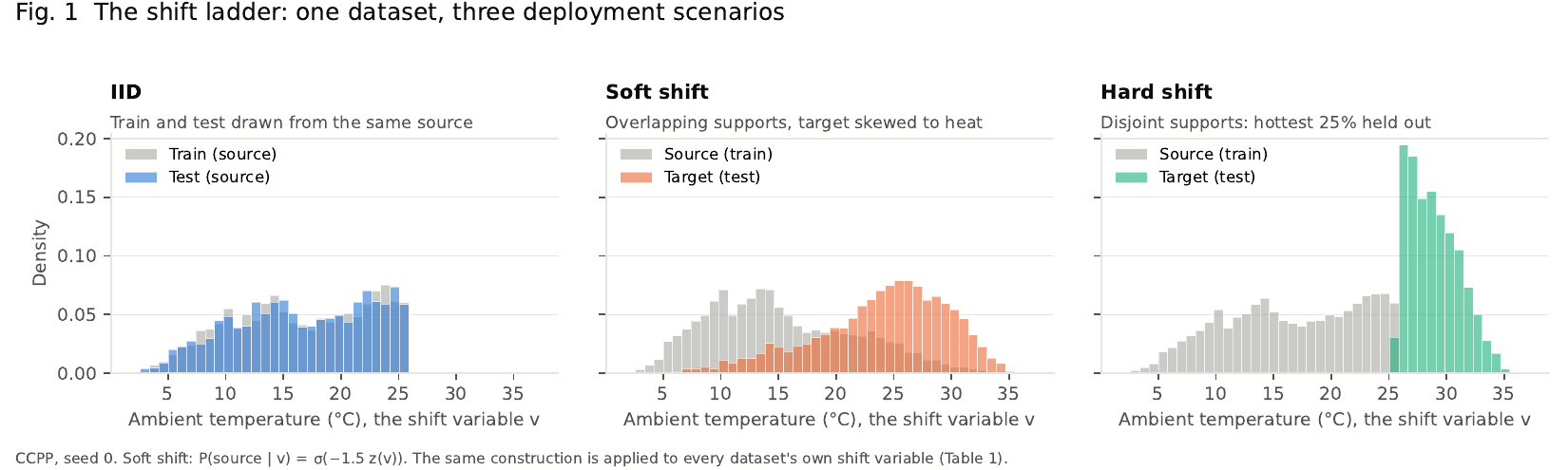}
\caption{The shift ladder on CCPP (seed~0). Soft shift: $P(\text{source}\mid v)=\sigma(-1.5\,z(v))$. Every dataset uses the same construction on its own shift variable (Table~\ref{tab:datasets}).}
\label{fig:ladder}
\end{figure}

We close this gap with a \emph{shift ladder}: three increasingly severe ways of separating training from deployment data, applied to the same datasets along a physically meaningful variable (Fig.~\ref{fig:ladder}). The first rung is the familiar random split. The second is a soft covariate shift, in which deployment is skewed towards one end of the shift variable but still overlaps the training data. The third is a hard shift. On five datasets it is an extrapolation, with no overlap at all: concrete older than any tested specimen, glazing beyond any simulated design, ambient temperatures, Froude numbers and ship speeds above the training range. On the other two it is a change of regime: a new fault mode in turbofan engines, and a sensor deployment that crosses from autumn into winter. The rungs sit on different sides of the assumptions behind weighted conformal prediction \citep{tibshirani2019conformal}. The soft rung is the setting that method is designed for; the hard rung is one in which no method that learns only from source data can guarantee coverage. The ladder therefore separates what methods fail to do from what cannot be done.

Along this ladder we evaluate 12 methods from five families on seven engineering datasets, with every method tuned on source data only, as a practitioner would have to. The headline is simple. In distribution, the conformal methods cover at least 0.946 on every dataset. Under hard shift, mean coverage falls to between 0.40 and 0.75 for the 11 methods that always return an interval, and no systematic difference in method ranks is detected. The one method that remains near nominal on the points it answers, weighted conformal prediction, does so by abstaining on 98\% of points on the extrapolation datasets, a legitimate safety response when the calibration data say nothing about the target.

Our contributions are:
\begin{enumerate}
\item\label{contrib:protocol} \textbf{A shift-ladder protocol for engineering regression.} It covers seven public datasets with physically meaningful shift variables, including one real field deployment over time. Code, results and a notebook that reproduces every table and figure will be released.
\item\label{contrib:joint} \textbf{A joint evaluation of Bayesian, ensemble, distributional, quantile and conformal methods along the ladder.} It includes weighted conformal prediction and uses nested hyperparameter tuning on source data only. To our knowledge, no published comparison covers these families together under graded, physically meaningful shift in engineering regression.
\item\label{contrib:marginal} \textbf{Evidence that marginal coverage hides failure.} Worst-subgroup coverage falls 0.05--0.10 below the mean under soft shift. GP intervals reach full coverage by becoming twice as wide as the target range. Coverage loss tracks the growth of point error (Spearman $\rho = 0.74$).
\item\label{contrib:mechanisms} \textbf{Three mechanisms of failure.} Tree-based intervals freeze beyond the training range; GP coverage is bought with width, at a cost set by the kernel; and ensembles rank their errors correctly but under-scale them.
\item\label{contrib:audit} \textbf{An audit of weighted conformal prediction with estimated weights.} Its abstention is driven by small effective calibration samples, not by weight-estimation error: exact weights abstain as often. It cannot detect conditional shift, and the few predictions it makes under extrapolation can still be wrong.
\item\label{contrib:tuning} \textbf{Evidence that tuning on source data does not transfer to hard shift.} Gains found on source validation predict improvement in distribution and under soft shift but not under hard shift ($\rho = 0.04$). The GP kernel chosen on source data moves hard-shift coverage by up to 0.47 in either direction.
\end{enumerate}

The rest of the paper is organised as follows. Section~\ref{sec:related} reviews related work. Section~\ref{sec:setting} defines the shift ladder and what each rung permits in theory. Section~\ref{sec:design} describes the datasets, methods, tuning protocol and metrics. Section~\ref{sec:results} presents the results, Section~\ref{sec:discussion} discusses their implications and limits, and Section~\ref{sec:conclusion} concludes.

\section{Related work}\label{sec:related}
 
UQ, distribution-free calibration and robustness to distribution shift have become central topics in machine learning research. In safety-critical engineering applications such as structural health monitoring, aerospace surrogate modelling and reliability engineering, predictive models must deliver not only accurate point predictions but also reliable, statistically valid bounds on their predictive uncertainty. This section reviews UQ methods for engineering regression (Section~\ref{sec:related-uq}), comparative benchmark studies (Section~\ref{sec:related-comparative}), conformal prediction under distribution shift (Section~\ref{sec:related-cp}), selective prediction (Section~\ref{sec:related-selective}) and model selection under shift (Section~\ref{sec:related-selection}), and closes with a gap analysis that positions this work (Section~\ref{sec:related-gap}).
 
\subsection{UQ for engineering regression}\label{sec:related-uq}
 
UQ in engineering regression generally addresses two distinct sources of error: aleatoric uncertainty (irreducible stochastic noise inherent to the physical data-generating process) and epistemic uncertainty (reducible uncertainty arising from limited training data, model misspecification or poorly chosen parameters). Surrogates used in structural reliability analysis, fatigue-life estimation and finite-element emulation have historically relied on Gaussian process regression (GPR) \citep{rasmussen2006gaussian}. GPs yield probabilistic prediction intervals through their kernel covariance function; however, exact inference scales as $\mathcal{O}(N^3)$ in the number of training points $N$, which restricts direct use on large datasets unless sparse or variational approximations are introduced \citep{titsias2009variational}.
 
To scale uncertainty estimation to complex, high-dimensional nonlinear functions, Bayesian neural networks (BNNs) place prior distributions over the network weights and approximate the weight posterior by mean-field variational inference \citep{blundell2015weight} or by Markov chain Monte Carlo sampling, including Hamiltonian Monte Carlo. While theoretically grounded, full Bayesian inference remains computationally demanding and sensitive to the choice of prior. Monte Carlo dropout (MC Dropout) approximates Bayesian inference at lower computational cost by keeping dropout active at test time to sample predictive distributions \citep{gal2016dropout}, although it often produces overconfident prediction intervals in regions of the input space not seen during training.
 
Deep ensembles, constructed by training several networks from different weight initialisations with proper scoring rules, such as the Gaussian negative log-likelihood with a homoscedastic or heteroscedastic variance \citep{lakshminarayanan2017simple}, have emerged as a strong empirical baseline for capturing epistemic uncertainty \citep{ovadia2019can}. SWA-Gaussian (SWAG) \citep{maddox2019simple} and MultiSWAG \citep{wilson2020bayesian} combine local approximations of the weight posterior with ensembling. Complementing parametric and Bayesian techniques, natural gradient boosting (NGBoost) \citep{duan2020ngboost} estimates the parameters of a conditional distribution directly, and quantile regression \citep{koenker1978regression} estimates chosen conditional quantiles without a parametric likelihood. These methods are increasingly used in reliability, condition monitoring and prognostics. However, intervals calibrated on historical operating conditions need not remain calibrated when assets, environments or operating regimes change \citep{ovadia2019can,barber2023conformal}.
 
\subsection{Comparative UQ studies}\label{sec:related-comparative}
 
Despite the proliferation of UQ methods, benchmarks that systematically compare their calibration and prediction-interval performance remain scarce.
 
The closest prior study to our work is the comparative study by \citet{dewolf2023valid}. They evaluated four classes of interval estimators (Bayesian methods, ensemble methods, direct interval estimation methods and conformal prediction) across a range of regression benchmarks. They found that the performance of uncalibrated methods fluctuated strongly across datasets, which they attributed to violations of assumptions inherent to some classes of methods, and that conformal post-hoc calibration restored nominal coverage. Their evaluation, however, was confined to repeated random train--test splits, in which training and test data are independent and identically distributed (IID). They noted that autocorrelated, time-series data violate the exchangeability on which these methods rest, but did not evaluate distribution shift.
 
In specialised domains, \citet{tran2020methods} compared UQ methods for materials property prediction, analysing calibration, sharpness and interval width on materials-property regression tasks. Similarly, \citet{psaros2023uncertainty} conducted an extensive comparative study for scientific machine learning (SciML). They integrated UQ methods with physics-informed neural networks (PINNs), stochastic partial differential equations (SPDEs) and neural operators (DeepONets), and introduced the open-source \texttt{NeuralUQ} library. Their work highlighted the sensitivity of epistemic uncertainty to network overparameterisation and hyperparameter selection, but focused on physics-constrained forward and inverse problems rather than on covariate shift in tabular engineering data.
 
Most recently, \citet{make2026tabular} examined the calibration of tabular classifiers under synthetic shift, applying Gaussian feature noise at graded intensities to benchmark datasets, and included split conformal prediction. Like earlier classification-centred studies of dataset shift \citep{ovadia2019can,rabanser2019failing}, it addresses discrete class labels rather than continuous regression targets, and so does not assess prediction-interval coverage and sharpness or abstention.
 
\subsection{Conformal prediction under shift}\label{sec:related-cp}
 
Conformal prediction (CP) provides a distribution-free framework for turning point predictions or heuristic interval estimators into prediction intervals with finite-sample coverage guarantees \citep{vovk2005algorithmic}. In inductive (split) conformal prediction (ICP), the data are partitioned into a proper training set $\mathcal{T}$ and a calibration set $\mathcal{V}$. Given a nonconformity measure $A(x,y)$, such as the absolute residual $|y-\hat y(x)|$ or, in conformalised quantile regression (CQR) \citep{romano2019conformalized}, the interval deviation $\max\big(\hat l(x)-y,\; y-\hat u(x)\big)$, ICP computes an empirical quantile of the nonconformity scores over $\mathcal{V}$ and constructs, for a new input $x^*$, a prediction interval $\hat C(x^*)$ that satisfies
\begin{equation}\label{eq:cp-marginal}
\Pr\big(y^* \in \hat C(x^*)\big) \ge 1-\alpha .
\end{equation}
 
This marginal guarantee relies on the exchangeability of calibration and test points. Under covariate shift, where the input distribution changes, $P_S(X)\neq P_T(X)$, while the conditional distribution of the target is unchanged, $P_S(Y\mid X)=P_T(Y\mid X)$, exchangeability is violated and unweighted conformal intervals can under-cover.
 
To restore validity under covariate shift, \citet{tibshirani2019conformal} introduced weighted conformal prediction. Reweighting the calibration scores by the likelihood ratio (importance weight) $w(x)=\mathrm{d}P_T(x)/\mathrm{d}P_S(x)$, they proved that the weighted empirical quantile gives valid coverage under the target covariate distribution when the ratio is known. Building on this, \citet{barber2023conformal} bounded the coverage loss of conformal methods beyond exchangeability by a weighted total-variation distance, a bound that applies when the shift or temporal drift cannot be specified exactly.
 
Weighted conformal intervals require an estimate of the target-to-source density ratio $w(x)$. Direct density-ratio estimators such as unconstrained least-squares importance fitting (uLSIF) and the Kullback--Leibler importance estimation procedure (KLIEP), reviewed by \citet{sugiyama2012density}, avoid estimating the high-dimensional source and target densities separately. The same framework covers probabilistic classification, in which a classifier trained to separate source from target inputs yields the ratio; this is the estimator used by Weighted CQR in this study (Section~\ref{sec:methods}).
 
\subsection{Selective prediction and abstention}\label{sec:related-selective}
 
In high-stakes engineering applications, a coverage guarantee is of limited use if the prediction intervals become too wide to inform a decision under shift. Under such conditions the system needs a reliable abstention mechanism, or reject option.
 
Selective prediction, formalised for deep networks by \citet{geifman2017selective}, introduces a selection function $g(x)\in\{0,1\}$ that determines whether a predictive model $f(x)$ outputs a prediction ($g(x)=1$) or abstains ($g(x)=0$). The objective is to optimise the trade-off between selective coverage (the share of inputs answered) and risk, keeping the risk on accepted inputs below a target tolerance $r^*$ with high probability.
 
To extend finite-sample guarantees beyond coverage, \citet{angelopoulos2024conformal} developed conformal risk control (CRC), which controls the expected value of any bounded loss $\ell$ that is monotone in a tuning parameter $\lambda$, for example a miscoverage loss or a bounded task loss. CRC chooses $\hat\lambda$ on calibration data so that
\begin{equation}\label{eq:crc}
\mathbb{E}\big[\ell\big(\hat C_{\hat\lambda}(X), Y\big)\big] \le \alpha
\end{equation}
in finite samples, with no distributional assumption beyond exchangeability. Applied to a loss that penalises abstention as well as error, it offers a formal bridge between conformal uncertainty estimation and selective prediction, a direction we return to in Section~\ref{sec:open}.
 
\subsection{Model selection under shift}\label{sec:related-selection}
 
An often overlooked part of evaluating distribution-robust machine-learning systems is the protocol for hyperparameter tuning and model selection. Standard $K$-fold cross-validation on source training data is biased under covariate shift, and configurations optimised on source data can be overfitted or overconfident when evaluated on the shifted target distribution.
 
To obtain unbiased validation under covariate shift, \citet{sugiyama2007covariate} proposed importance-weighted cross-validation (IWCV), which reweights the validation loss of each point by the density ratio $w(x_i)$ and gives an asymptotically unbiased estimator of the target generalisation error. In domain generalisation, \citet{gulrajani2021search} showed that the model-selection protocol can dominate reported benchmark performance: under a consistent protocol, carefully tuned empirical risk minimisation matched most proposed algorithms, and they argued that an algorithm is incomplete without its model-selection method. This frames contribution~\ref{contrib:tuning} in Section~\ref{sec:intro}: we tune every method on source data only, as a practitioner must before deployment, and measure whether the gains carry over along the shift ladder.
 
\subsection{Summary gap analysis}\label{sec:related-gap}
 
Taken together, the five strands above leave a specific gap. Comparative studies of regression UQ \citep{dewolf2023valid,tran2020methods,psaros2023uncertainty} evaluate methods on random hold-out or standard cross-validation splits, so they measure calibration only where training and test data are exchangeable. Studies that do introduce graded shift \citep{ovadia2019can,make2026tabular} address classification, with synthetic corruptions or feature noise, rather than regression targets under physically meaningful shift. Weighted conformal prediction has a clear theory under covariate shift \citep{tibshirani2019conformal}, but its behaviour with estimated weights near or beyond the limits of that theory (small effective samples, non-overlapping support, or a change in $P(Y\mid X)$) has not been audited on engineering data. Selective prediction and conformal risk control provide a formal language for abstention \citep{geifman2017selective,angelopoulos2024conformal}, yet none of the comparative studies reviewed above reports abstention or subgroup coverage. Finally, the model-selection literature shows that the tuning protocol can dominate reported performance \citep{gulrajani2021search}, whereas those comparisons tune on random hold-out splits or use fixed defaults, so whether source-domain tuning helps under shift is untested.
 
This study addresses these gaps in six respects:
\begin{enumerate}
\item \textbf{Target space.} We evaluate continuous engineering regression targets rather than discrete class labels.
\item \textbf{Domain data.} We use seven public engineering datasets with physically meaningful shift variables, drawn from simulators, laboratory tests and field measurements, including one field deployment over time.
\item \textbf{Graded distribution shift.} Rather than random splits alone, or a single in-domain/out-of-domain boundary, we evaluate every method along a three-rung shift ladder (IID, soft shift and hard shift) defined on a physically meaningful shift variable, so that calibration can be tracked as the shift becomes more severe.
\item \textbf{Subgroup coverage.} We report worst-subgroup coverage alongside marginal coverage, sharpness and point error, to expose failures that averages hide.
\item \textbf{Weighted conformal prediction and abstention.} We include weighted conformal prediction, which abstains by returning an infinite interval when its calibration data cannot support a finite one, and we audit when it abstains, what drives its abstention and whether the predictions it does make are covered.
\item \textbf{Tuning protocol.} Every method is tuned by nested search on source data only, as a practitioner must before deployment, and we measure whether the gains carry over along the ladder.
\end{enumerate}

\section{Problem setting and the shift ladder}\label{sec:setting}

\subsection{Notation and coverage}\label{sec:notation}

Let $X\in\mathbb{R}^d$ be the input and $Y\in\mathbb{R}$ the target. Labelled data come from a source distribution $P_S(X,Y)$, and the model is deployed on a target distribution $P_T(X,Y)$. A UQ method returns, for each input $x$, a prediction interval $\hat C(x)=[l(x),u(x)]$ at miscoverage level $\alpha=0.05$. Its marginal coverage on the target, the prediction interval coverage probability (PICP), is
\begin{equation}\label{eq:picp-t}
\PICP_T = \Pr_{(X,Y)\sim P_T}\big(Y \in \hat C(X)\big).
\end{equation}

Marginal coverage can sit at 0.95 while some operating regimes are covered far less often. We therefore also report subgroup coverage. For a partition of the input space into groups $G = 1,\dots,K$, defined per dataset in Table~\ref{tab:datasets}, the worst-group coverage is
\begin{equation}\label{eq:picp-min}
\PICP_{\min} = \min_{g}\; \Pr_{P_T}\big(Y \in \hat C(X) \mid G = g\big).
\end{equation}

Conditional coverage at every $x$ cannot be guaranteed by any finite-sample method, so the worst-group figure is the practical proxy for it. A method \emph{abstains} at $x$ when it returns $\hat C(x)=\mathbb{R}$. We call a method \emph{selective} if it can abstain and \emph{non-selective} otherwise; only Weighted CQR is selective, and the other 11 methods always return a finite interval. For a selective method, coverage is computed only over the points where it answers. It estimates $\Pr\{Y\in\hat C(X)\mid \text{answer}\}$, not unconditional deployment coverage, so it is always reported with the abstention rate and the number of answered points.

Split conformal prediction guarantees $\Pr(Y\in\hat C(X))\ge 1-\alpha$ whenever the calibration and test points are exchangeable. The shift ladder is built to control, step by step, how far that assumption is violated.

\subsection{Three rungs}\label{sec:rungs}

Each dataset has a physically meaningful shift variable $v(x)$: curing age, glazing area, ambient temperature, Froude number, ship speed or, for Air Quality, calendar time. For Concrete we use log age. In every dataset except Air Quality, $v$ is also a model input. Let $z(v)$ be $v$ standardised over the full dataset, and $\tau$ a threshold on $v$ fixed per dataset (Table~\ref{tab:datasets}). Fig.~\ref{fig:ladder} shows the three rungs on the combined cycle power plant (CCPP) dataset.

\textbf{Rung 1, IID.} The source domain is $\{v\le\tau\}$. It is split at random into 70\% for training and 30\% for testing. Training and test points are exchangeable by construction, so this rung reproduces the random-split protocol of earlier benchmarks.

\textbf{Rung 2, soft covariate shift.} Each row is assigned to the source independently, with probability
\begin{equation}\label{eq:soft}
\pi(v) = \Pr(\text{source}\mid v) = \sigma\big(-k\,z(v)\big), \qquad k = 1.5,
\end{equation}
and every row not selected forms the target. Both domains cover the full range of $v$, but the target is concentrated at high $v$. Selection depends on $x$ only, so $P(Y\mid X)$ is the same in both domains, which makes this a pure covariate shift. The true density ratio has a closed form:
\begin{equation}\label{eq:ratio}
w(x) = \frac{p_T(x)}{p_S(x)} \;\propto\; \frac{1-\pi(v)}{\pi(v)} = \exp\big(k\,z(v)\big),
\end{equation}
which is bounded on the observed range of $v$. For the Commercial Modular Aero-Propulsion System Simulation (C-MAPSS) data there is no scalar shift variable, so the soft rung is built differently. The source is all FD001 engines plus a random 20\% of FD003 engines, and the target is the remaining FD003 engines. The source has therefore seen the new fault mode, but only rarely. Selection here acts on fault mode, which is not an input, so this rung is not a pure covariate shift either. For Air Quality, selection acts on time, which is not an input. The soft rung is then a covariate shift in $(X, v)$ jointly but not in $X$ alone: $P(Y\mid X)$ can differ between the domains through seasonal effects and sensor drift, and the closed-form ratio $\exp(k\,z(v))$ is a weight on time, not on $x$.

\textbf{Rung 3, hard shift.} The source is $\{v\le\tau\}$ and the target is $\{v>\tau\}$. The rung takes two forms. For five datasets (Concrete, Energy, CCPP, Yacht, Naval), $v$ is an input and its supports do not overlap, so every target prediction is an \emph{extrapolation}. For the other two, the covariates overlap and the shift is in $P(Y\mid X)$. For C-MAPSS, the source is every FD001 engine (one fault mode, high-pressure compressor (HPC) degradation) and the target is every FD003 engine (compressor plus fan degradation); both share one operating condition, so their sensor distributions overlap widely. For Air Quality, the source is every hour before 1 December 2004 and the target every hour after it. At most 1.6\% of target points fall outside the source range on any single feature, but winter conditions and sensor ageing change the mapping from sensor response to CO. We call these two \emph{conditional} (deployment) shifts and analyse them separately from the five extrapolation datasets where the distinction matters. ``Hard shift'' is therefore an operational severity category: the target regime has little or no representation in the source. We do not assume that all seven hard-shift targets follow the same formal distribution-shift model.

\subsection{What each rung permits in theory}\label{sec:theory}

The rungs are chosen because they sit on different sides of the assumptions behind weighted conformal prediction \citep{tibshirani2019conformal}. That method keeps its coverage guarantee under covariate shift if $P_T$ is absolutely continuous with respect to $P_S$ and the density ratio $w(x)$ is known. Under the soft rung both conditions hold, except that $w$ must be estimated. The rung therefore tests how much of the guarantee survives estimation error and a small effective calibration sample. Because the true $w$ is known in closed form, an oracle-weight variant can isolate that estimation error (Section~\ref{sec:res-wcqr}). For Air Quality the soft rung selects on time, which weights estimated on $x$ cannot recover; oracle weights on time can.

Under the five extrapolation rungs, $P_T$ puts mass where $P_S$ has none. The density ratio is then unbounded, and no calibration-based method can guarantee coverage. The source calibration set carries no information about residuals in the target region. The correct behaviour of weighted conformal prediction is to let the test-point weight dominate the weighted quantile, which returns an infinite interval: it abstains. Methods without a coverage guarantee behave according to their inductive bias off the support. A stationary GP reverts to its prior. Tree ensembles predict a constant beyond the last split. Networks with rectified linear unit (ReLU) activations extrapolate linearly with a variance head that was never trained there.

Conditional shift, as in C-MAPSS and Air Quality, is harder to detect than extrapolation. The covariates overlap, so a density-ratio classifier assigns finite weights and weighted conformal prediction answers rather than abstaining. Its guarantee is void, because reweighting $X$ cannot correct a change in $P(Y\mid X)$. The coverage-gap bounds of \citet{barber2023conformal} make this explicit: the gap grows with the distance between calibration and test distributions, which reweighting on $X$ does not reduce. Coverage on this rung is therefore an empirical outcome, not a guaranteed one, and it is reported as such.

\section{Experimental design}\label{sec:design}

\subsection{Datasets}\label{sec:datasets}

We use seven public engineering regression datasets. In each, the shift variable corresponds to a real deployment decision: using a model beyond the curing ages, designs, weather, speeds, fault modes or calendar period it was trained on. Five are the versions from the UCI Machine Learning Repository used by \citet{gal2016dropout}. The sixth is NASA C-MAPSS, for which we use the FD001 and FD003 training files. The seventh is the UCI Air Quality dataset \citep{devito2008field}: hourly responses of a five-channel metal-oxide sensor device co-located with a certified reference analyser on a city road in Italy, March 2004 to April 2005. It is the only dataset of real field measurements taken over time. The target is the analyser's CO concentration; the inputs are the five sensor channels plus temperature, relative humidity and absolute humidity. The other reference-analyser columns are excluded, and hours with any missing value (coded $-200$) are dropped, leaving 7,344 of 9,357 hours. For C-MAPSS, remaining useful life (RUL) is capped at 125 cycles (a piecewise-linear target). Each of 14 informative sensors is smoothed with a 10-cycle trailing mean per engine, and the cycle count is added as a feature. For Naval, two constant columns are removed.

\begin{table*}[width=\textwidth,cols=9,pos=t]
\centering
\caption{Datasets, shift definitions and the grouping used for worst-group coverage.}
\label{tab:datasets}
\scriptsize
\setlength{\tabcolsep}{3pt}
\begin{tabularx}{\textwidth}{@{}l r r L L L >{\raggedleft\arraybackslash}p{0.75cm} L L@{}}
\toprule
Dataset & Rows & Feat. & Target & Shift variable $v$ & Hard shift: source $\to$ target & Target rows & Groups for $\PICP_{\min}$ & Deployment scenario \\
\midrule
Concrete & 1,030 & 8 & Compressive strength (MPa) & Curing age (log days) & age $\le$ 28\,d $\to$ age $>$ 28\,d & 281 & 4 age quantile bands & Predicting long-term strength from early-age tests \\
Energy & 768 & 8 & Heating load (kWh/m$^2$) & Glazing area ratio & glazing $\le$ 25\% $\to$ 40\% & 240 & 4 relative-compactness bands & Assessing a more glazed design than any simulated \\
CCPP & 9,568 & 4 & Net electrical output (MW) & Ambient temperature (\textdegree C) & coolest 75\% of hours $\to$ hottest 25\% & 2,388 & 5 relative-humidity bands & Plant output during a heatwave \\
Yacht & 308 & 6 & Residuary resistance per unit weight & Froude number & Fr $\le$ 0.35 $\to$ Fr $>$ 0.35 & 88 & 3 Froude bands & Hull resistance at higher speed than tank-tested \\
Naval & 11,934 & 14 & Gas-turbine (GT) compressor decay coefficient & Ship speed (knots) & $\le$ 21\,kn $\to$ 24--27\,kn & 2,652 & 3 speed bands & Condition monitoring at top-speed regimes \\
C-MAPSS & 45,351 & 15 & RUL, capped at 125 cycles & Fault mode & FD001 (HPC fault) $\to$ FD003 (HPC + fan faults) & 24,720 & 5 RUL bands (0--25 \dots\ 100--125) & Prognostics under an unseen fault mode \\
Air Quality & 7,344 & 8 & Reference CO concentration (mg/m$^3$) & Calendar time (not an input) & before 1 Dec 2004 $\to$ from 1 Dec 2004 & 2,573 & 4 ambient-temperature bands & Keeping a low-cost sensor calibration in service through its first winter \\
\bottomrule
\end{tabularx}
\end{table*}

Subgroups are quantile bands computed on each evaluation set. A group enters $\PICP_{\min}$ only if it holds at least 15 test points. For Concrete, Yacht and Naval, the grouping variable is the shift variable itself, so $\PICP_{\min}$ measures how coverage decays with distance from the source. For Energy and CCPP it is a second physical variable. For Air Quality it is ambient temperature, to which metal-oxide sensors are cross-sensitive. For C-MAPSS it is the RUL band, which isolates coverage near end of life, the region where maintenance decisions are made.

\subsection{Methods}\label{sec:methods}

We compare 12 methods from five families (Table~\ref{tab:methods}). Each captures a different source of uncertainty, and that source predicts how the method should behave under shift. Methods that model epistemic uncertainty should widen their intervals away from the data. Methods that model only aleatoric noise have no reason to. Conformal methods inherit whatever the base model does, then rescale it using a calibration set drawn from the source.

The paragraphs below describe how each family builds a 95\% interval and what that construction implies away from the training data. Full equations are in Appendix~A of the supplementary material. All methods are fitted on standardised inputs and targets.

\textbf{Bayesian methods.} \emph{GPR} \citep{rasmussen2006gaussian} places a prior over functions defined by a kernel and returns a Gaussian predictive distribution: a mean and a variance that combines uncertainty about the function with observation noise. The interval is $\mu\pm1.96\sigma$. For a stationary kernel the function uncertainty grows with distance from the training inputs. Far from the data, the mean returns to the prior mean and the variance to the prior variance. A GP therefore widens its intervals when it extrapolates, but by an amount set by the kernel's length-scales and family, not by the data in the new region. The \emph{BNN} learns a Gaussian distribution over every weight by variational inference (VI) \citep{blundell2015weight}; \emph{MC Dropout} approximates the same idea by keeping dropout active at prediction time \citep{gal2016dropout}. In both, each network outputs a mean and a variance, and 100 sampled networks are combined by moment matching. The predictive variance is the spread of the sampled means (epistemic) plus the average predicted variance (aleatoric).

\textbf{Ensemble.} The \emph{deep ensemble} \citep{lakshminarayanan2017simple} trains five networks with the same architecture from different random initialisations and data orders, each with a mean-and-variance output trained on the Gaussian likelihood. The interval comes from the moment-matched mixture, so its width grows where members disagree. With ReLU activations each member extrapolates linearly, and members can either diverge or agree beyond the data. Their disagreement therefore signals novelty only when the members happen to extrapolate differently.

\textbf{Distributional.} \emph{NGBoost} \citep{duan2020ngboost} fits the mean and log standard deviation of a Normal distribution with gradient-boosted regression trees, using the natural gradient. It models how noise varies across the input space (aleatoric uncertainty) but has no term for uncertainty about the function itself. As with any tree model, its predicted mean and spread are constant beyond the last split of each tree.

\textbf{Quantile methods.} These predict the 2.5\% and 97.5\% conditional quantiles directly, without assuming a distribution. \emph{Quantile regression forests} (Quantile RF) \citep{meinshausen2006quantile} keep every training response in the leaves and read the quantiles off the responses that share leaves with the query point. The \emph{LightGBM quantile} model \citep{ke2017lightgbm} fits a separate boosted model for each quantile with the pinball (quantile) loss. The \emph{quantile regression network} (QRNN) \citep{taylor2000quantile} is one network with three outputs trained on the same loss. All three describe the spread of the data around the regression function. They do not describe how uncertain the function is, and the two tree-based methods return the quantiles of the nearest training leaves for any input beyond the training range.

\textbf{Conformal methods.} These wrap a base model and calibrate it on held-out data. The base model is fitted on 80\% of the training rows. A \emph{nonconformity score} is computed on the remaining 20\%, and the interval is widened by the empirical $(1-\alpha)$ quantile of those scores. When calibration and test points are exchangeable, this guarantees coverage of at least 95\% regardless of the base model \citep{vovk2005algorithmic}. \emph{Split CP} uses the absolute residual of a point model, which gives intervals of constant width. \emph{CQR} \citep{romano2019conformalized} uses the distance by which a point falls outside a base quantile band, so it inherits the band's adaptive width. \emph{Conformal GPR} divides the residual by the GP's predictive standard deviation, so its width scales with the GP's own uncertainty. \emph{Weighted CQR} \citep{tibshirani2019conformal} reweights the calibration scores by an estimate of the density ratio between test and calibration inputs. Calibration points that resemble the test point count for more, and the test point's own weight is added to the total. When that weight is so large that no finite score reaches the required 95\% of weighted mass, the interval is infinite: the method abstains. Conformal methods inherit the base model's behaviour off the support. Their guarantee rests on exchangeability, which every shift in this benchmark violates.

\begin{table*}[width=\textwidth,cols=5,pos=t]
\centering
\caption{Methods, the uncertainty each represents and their settings. The last column gives the search space; the first value listed is the default, used in the fixed-setting run. Conformal methods inherit their base model's tuned configuration (Section~\ref{sec:protocol}).}
\label{tab:methods}
\scriptsize
\setlength{\tabcolsep}{3pt}
\begin{tabularx}{\textwidth}{@{}>{\raggedright\arraybackslash}p{1.6cm} >{\raggedright\arraybackslash}p{1.85cm} >{\raggedright\arraybackslash}p{2.3cm} >{\raggedright\arraybackslash}p{2.6cm} L@{}}
\toprule
Method & Family & Uncertainty represented & Interval & Fixed settings; search space (default first) \\
\midrule
GPR & Bayesian & Epistemic + homoscedastic aleatoric & $\mu\pm1.96\sigma$ & Type-II maximum likelihood (GPy); exact up to 2,000 rows, else sparse with 200 $k$-means inducing points. Automatic relevance determination (ARD) kernel $\in$ \{radial basis function (RBF), Mat\'ern-5/2, Mat\'ern-3/2\} \\
BNN (VI) & Bayesian & Epistemic (weights) + heteroscedastic aleatoric & $\mu\pm1.96\sigma$, moment-matched over 100 samples & Mean-field Gaussian posterior, Bayes-by-backprop, up to 600 epochs, patience 40. Width $\in$ \{64, 32, 128\}; prior sd $\in$ \{1, 0.1, 0.5\}; learning rate $\in$ \{$10^{-3}$, $3\cdot10^{-3}$\} \\
MC Dropout & Bayesian (approx.) & Epistemic + heteroscedastic aleatoric & As BNN & 100 stochastic passes, patience 30. Dropout $\in$ \{0.1, 0.05, 0.2, 0.3\}; width $\in$ \{64, 128\}; weight decay $\in$ \{$10^{-5}$, $10^{-4}$\}; learning rate $\in$ \{$10^{-3}$, $3\cdot10^{-3}$\} \\
Deep Ensemble & Ensemble & Member disagreement + heteroscedastic aleatoric & Gaussian mixture, moment-matched & 5 members with Gaussian negative log-likelihood heads, patience 30. Width $\in$ \{64, 32, 128\}; learning rate $\in$ \{$10^{-3}$, $3\cdot10^{-3}$\}; weight decay $\in$ \{0, $10^{-4}$\} \\
NGBoost & Distributional & Heteroscedastic aleatoric & $\mu\pm1.96\sigma$ & Normal distribution, up to 1,000 trees, early stopping after 50 rounds. Learning rate $\in$ \{0.03, 0.01, 0.1\}; tree depth $\in$ \{3, 2, 4, 5\}; minibatch fraction $\in$ \{1, 0.5\} \\
Quantile RF & Quantile & Conditional quantiles (aleatoric) & Empirical 2.5\% and 97.5\% quantiles & 300 trees. Minimum leaf $\in$ \{5, 1, 10, 20, 40\}; features per split $\in$ \{all, half, $\sqrt{d}$\} \\
LightGBM Quantile & Quantile & Conditional quantiles (aleatoric) & Separate models at $\tau = 0.025$, 0.5, 0.975 & Row subsampling 0.8. Leaves $\in$ \{31, 7, 15, 63\}; learning rate $\in$ \{0.03, 0.01, 0.1\}; minimum child samples $\in$ \{10, 5, 20, 40\}; trees $\in$ \{500, 200, 1,000\} \\
QRNN & Quantile & Conditional quantiles (aleatoric) & Three-output network, pinball loss, outputs sorted & Patience 30. Width $\in$ \{64, 32, 128\}; learning rate $\in$ \{$10^{-3}$, $3\cdot10^{-3}$\}; weight decay $\in$ \{0, $10^{-5}$, $10^{-4}$\} \\
Split CP & Conformal & Distribution-free, constant width & $\mu\pm\hat q$ on absolute residuals & LightGBM mean model; same grid as LightGBM Quantile, selected on validation mean squared error (MSE) \\
CQR & Conformal & Distribution-free, adaptive width & Quantile band widened by $\hat q$ \citep{romano2019conformalized} & LightGBM quantile base; inherits its tuned configuration \\
Conformal GPR & Conformal & Distribution-free, width scaled by GP $\sigma$ & $\mu\pm\hat q\sigma$ on normalised residuals & GPR base; inherits its tuned kernel \\
Weighted CQR & Conformal & Distribution-free under covariate shift & CQR with a weighted quantile; infinite interval when the quantile is unattainable \citep{tibshirani2019conformal} & Density ratio from degree-2 polynomial logistic regression ($C = 0.1$); shares the tuned CQR model and its calibration set \\
\bottomrule
\end{tabularx}
\end{table*}

\subsection{Protocol}\label{sec:protocol}

Every method is trained once per dataset, rung and seed on the source domain. It is then scored on the relevant test set. Inputs and targets are standardised using the training rows only, and predictions are mapped back to original units before scoring. For each rung, the source is split 70/30 into training and IID test rows. C-MAPSS is split by whole engine, so no engine contributes to both sides. The IID rung reported in Section~\ref{sec:results} uses the source from the hard-shift split. The soft-shift runs refit every model on their own source sample.

Training sets are capped at 3,000 rows and test sets at 2,000, sampled at random. The cap is binding for CCPP, Naval and C-MAPSS in both ladder modes, and for Air Quality in the hard mode. Under the hard rung, Concrete, Energy and Yacht train on 525, 370 and 154 rows respectively. Every configuration is repeated over five seeds (0--4), which vary the data split, weight initialisation and calibration split. This gives 70 runs (7 datasets $\times$ 2 ladder modes $\times$ 5 seeds), each scored on an IID and a shifted test set.

Conformal methods fit their base model on 80\% of the training rows and calibrate on the remaining 20\%, so their base models see less data than the non-conformal methods. For C-MAPSS, the calibration set is drawn as whole engines, because row-level calibration on correlated time series inflates exchangeability (Section~\ref{sec:res-grouped} measures the effect). Weighted CQR is transductive. Its density-ratio classifier separates the proper-training inputs from the unlabelled inputs of the evaluation set, and is fitted separately for the IID and shifted test sets. No target labels are used by any method.

\emph{Hyperparameters.} All neural networks use two hidden ReLU layers, Adam (batch size 128) and early stopping on a 10\% validation split. Every method is tuned inside every run, on the source training rows only. Those rows are split once more, 80/20, into inner-train and inner-validation rows; for C-MAPSS the split is by whole engine. Neither the target domain nor the IID test rows are used at any point, so the selection is one a practitioner could make before deployment. Each candidate configuration is fitted on inner-train and scored by its mean Winkler score (Section~\ref{sec:metrics}) on inner-validation, and the best is refitted on all training rows. The Winkler score is a proper score for central intervals and puts every family on one scale. The one exception is the point LightGBM behind Split CP, which is selected on validation MSE, because the conformal residual score needs only a good conditional mean.

Search is random over the grids in Table~\ref{tab:methods} \citep{bergstra2012random}: 8 candidates per method, 3 for GPR (all kernels) and 12 for the two LightGBM models. The default configuration is always one of the candidates, so tuning cannot do worse than the defaults on the inner criterion. Conformal methods are not searched separately: CQR and Weighted CQR inherit the tuned LightGBM-quantile configuration, Conformal GPR the tuned GPR kernel, and Split CP the tuned point LightGBM. To bound cost, inner-train is capped at 1,500 rows, which affects only the four capped datasets; the final refit uses all training rows. The fixed-setting run with the defaults is kept, and Section~\ref{sec:res-tuning} compares the two.

\subsection{Metrics}\label{sec:metrics}

All interval metrics are normalised by the target range $R$ of the full dataset (source and target together), so values are comparable across datasets. $R$ is fixed per dataset and is therefore the same on every rung. Because it uses target labels, it is a normaliser for retrospective evaluation, not a quantity available at deployment. For $n$ test points with intervals $[l_i,u_i]$ and point predictions $\mu_i$, the coverage and the prediction interval normalised average width (PINAW) are
\begin{equation}\label{eq:picp-pinaw}
\begin{aligned}
\PICP &= \tfrac{1}{n}\sum_i \mathbb{1}[l_i \le y_i \le u_i], \\
\mathrm{PINAW} &= \tfrac{1}{nR}\sum_i (u_i - l_i),
\end{aligned}
\end{equation}

and the Winkler score is

\begin{equation}\label{eq:winkler}
\begin{split}
\mathrm{Winkler} = {} & \tfrac{1}{nR}\sum_i \Big[(u_i - l_i) + \tfrac{2}{\alpha}(l_i - y_i)\,\mathbb{1}[y_i < l_i] \\
& + \tfrac{2}{\alpha}(y_i - u_i)\,\mathbb{1}[y_i > u_i]\Big].
\end{split}
\end{equation}

Lower Winkler scores are better. For a point inside its interval the score is simply the interval's width. A point outside adds $2/\alpha = 40$ times its distance from the nearer bound. A method therefore cannot score well by being narrow and often wrong, nor by being wide and always right.

Coverage alone rewards uninformative intervals: returning $\mathbb{R}$ everywhere achieves $\PICP = 1$. We therefore read coverage alongside sharpness (PINAW) and use the Winkler score, which is a proper scoring rule for central intervals, to rank methods. For the five Gaussian methods we also report the negative log-likelihood (NLL) and the closed-form continuous ranked probability score (CRPS), the latter normalised by $R$. NLL is reported per dataset only, because it depends on the target's units.

Point accuracy is measured by the normalised root-mean-square error (NRMSE) of the point prediction $\mu_i$, normalised by the same range $R$ so that it shares the scale of the interval metrics:
\begin{equation}\label{eq:nrmse}
\mathrm{NRMSE} = \frac{1}{R}\sqrt{\tfrac{1}{n}\textstyle\sum_i (y_i - \mu_i)^2}.
\end{equation}

NRMSE answers a different question from coverage. It says how far the predictions are from the truth; coverage says whether the intervals admit it. Section~\ref{sec:res-hard} shows that the two move together under shift but rank methods differently.

Two further diagnostics ask \emph{why} coverage fails. The first is worst-group coverage $\PICP_{\min}$ (Eq.~\eqref{eq:picp-min}). The second is the Spearman correlation between interval width and absolute error, $u_i-l_i$ against $|y_i-\mu_i|$. A method with high correlation but low coverage ranks risk correctly and misjudges its scale, which recalibration might repair. A method with low correlation cannot be repaired by rescaling. For Weighted CQR we report the abstention rate, coverage on the points it answers, and the effective calibration sample size (ESS)
\begin{equation}\label{eq:ess}
\mathrm{ESS} = \frac{\big(\sum_{j} w(x_j)\big)^2}{\sum_j w(x_j)^2}
\end{equation}
over calibration points $j$. A small ESS signals that a few calibration points carry the weighted quantile, and it predicts abstention.

\subsection{Statistical testing}\label{sec:stats}

To test whether coverage drops under shift, we apply a paired Wilcoxon signed-rank test per method to IID versus shifted PICP, over dataset--seed pairs ($n = 35$; $n = 20$ for Weighted CQR under hard shift, where it abstains on every point of three datasets). Each shifted result is paired with the IID result of the same fitted models, so the soft rung is paired with the soft-mode IID test set and the hard rung with the hard-mode one. The 24 tests (12 methods $\times$ 2 rungs) are Holm-corrected as one family. The 35 pairs are five seeds on each of seven datasets. Seeds are repeated computational realisations, not independent engineering domains, so these tests establish that a drop is consistent across seeds on these seven datasets; they do not test generalisation to other datasets. Claims about datasets rest on the per-dataset results (Fig.~\ref{fig:heatmaps}) and on effect sizes rather than on $p$-values.

To test whether methods differ in interval quality within a rung, we apply a Friedman test to dataset-level mean Winkler scores (7 datasets as blocks, 11 methods). Weighted CQR is excluded, because its Winkler score is defined only on the points where it answers. With seven blocks the test has limited power, so a non-significant result is read as no detectable difference, not as equivalence. No rung gives a significant Friedman result (Section~\ref{sec:results}), so no post-hoc critical-difference analysis is reported.

The effect of tuning is tested per method and rung with a paired Wilcoxon test on Winkler scores, tuned versus fixed settings, over the same 35 dataset--seed pairs; the 36 tests are Holm-corrected. Whether gains found on source validation carry over to a rung is measured by the Spearman correlation, across the 8 searched methods $\times$ 7 datasets $\times$ 5 seeds ($n = 280$ per rung; the IID rung uses the hard-mode runs), between the inner-validation Winkler change and the test-set Winkler change.

\section{Results}\label{sec:results}

\begin{figure}[pos=tbp]
\centering
\includegraphics[width=0.85\linewidth,trim=0 18 0 19,clip]{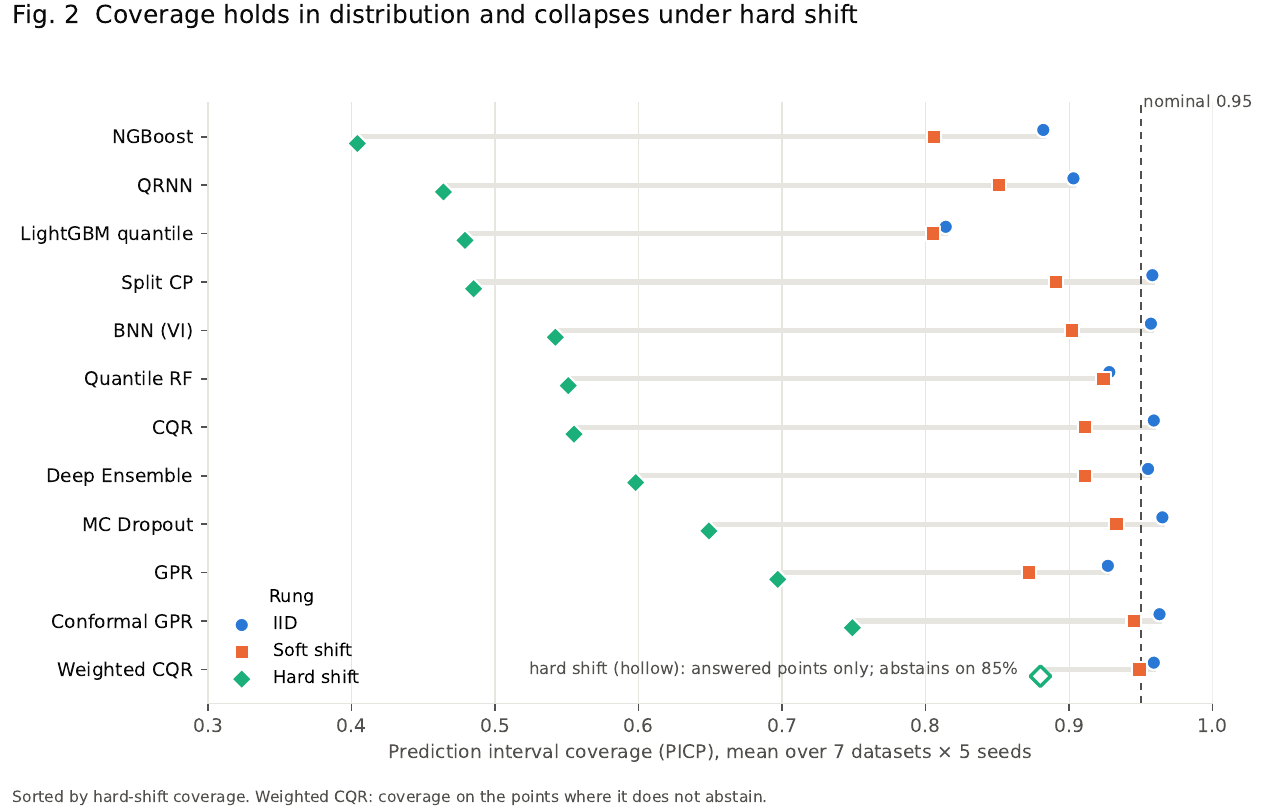}
\caption{Mean coverage by rung, sorted by hard-shift coverage. Weighted CQR: coverage on answered points.}
\label{fig:coverage-ladder}
\end{figure}

Coverage holds in distribution, erodes under soft shift and collapses under hard shift, and no non-selective method keeps nominal coverage on the last rung. Table~\ref{tab:summary} and Fig.~\ref{fig:coverage-ladder} summarise the ladder across the seven datasets; per-dataset values with standard deviations are in the supplement. All results in this section are from the tuned run. Section~\ref{sec:res-tuning} shows that the fixed-setting run leads to the same conclusions.

\begin{table*}[width=\textwidth,cols=10,pos=t]
\centering
\caption{Cross-dataset summary (tuned run; mean over 7 datasets of 5-seed means). $\PICP_{\min}$: worst-subgroup coverage. NRMSE: root-mean-square error divided by the target range; Weighted CQR shares CQR's point model. Winkler rank: mean rank among the 11 non-selective methods (1 = best). \textsuperscript{a}On answered points, over the four datasets where Weighted CQR answers any.}
\label{tab:summary}
\scriptsize
\setlength{\tabcolsep}{3pt}
\begin{tabular}{@{}llcccccccc@{}}
\toprule
 & & \multicolumn{4}{c}{PICP} & PINAW & \multicolumn{2}{c}{NRMSE} & Winkler \\
\cmidrule(lr){3-6}\cmidrule(lr){7-7}\cmidrule(lr){8-9}\cmidrule(l){10-10}
Method & Family & IID & Soft & Min, soft & Hard & Hard & IID & Hard & rank, hard \\
\midrule
GPR & Bayesian & 0.93 & 0.87 & 0.79 & 0.70 & 0.56 & 0.044 & 0.209 & 5.7 \\
BNN (VI) & Bayesian & 0.96 & 0.90 & 0.81 & 0.54 & 0.29 & 0.061 & 0.303 & 7.0 \\
MC Dropout & Bayesian & 0.96 & 0.93 & 0.85 & 0.65 & 0.31 & 0.052 & 0.229 & 4.9 \\
Deep Ensemble & Ensemble & 0.96 & 0.91 & 0.81 & 0.60 & 0.41 & 0.051 & 0.362 & 6.0 \\
NGBoost & Distributional & 0.88 & 0.81 & 0.72 & 0.40 & 0.15 & 0.050 & 0.185 & 7.9 \\
Quantile RF & Quantile & 0.93 & 0.92 & 0.87 & 0.55 & 0.26 & 0.054 & 0.186 & 4.4 \\
LightGBM Quantile & Quantile & 0.81 & 0.81 & 0.72 & 0.48 & 0.25 & 0.050 & 0.190 & 7.1 \\
QRNN & Quantile & 0.90 & 0.85 & 0.77 & 0.46 & 0.27 & 0.050 & 0.391 & 7.0 \\
Split CP & Conformal & 0.96 & 0.89 & 0.79 & 0.48 & 0.22 & 0.048 & 0.180 & 6.0 \\
CQR & Conformal & 0.96 & 0.91 & 0.83 & 0.56 & 0.31 & 0.052 & 0.190 & 5.0 \\
Conformal GPR & Conformal & 0.96 & 0.94 & 0.90 & 0.75 & 0.67 & 0.046 & 0.214 & 5.0 \\
Weighted CQR & Conformal & 0.96 & 0.95 & 0.91 & 0.88\textsuperscript{a} & 0.36 & 0.052 & 0.190 & -- \\
\bottomrule
\end{tabular}
\end{table*}

\subsection{In distribution, conformal methods are calibrated everywhere and the neural Bayesian methods nearly so}\label{sec:res-iid}

\begin{figure}[pos=tbp]
\centering
\includegraphics[width=\linewidth,trim=0 24 0 19,clip]{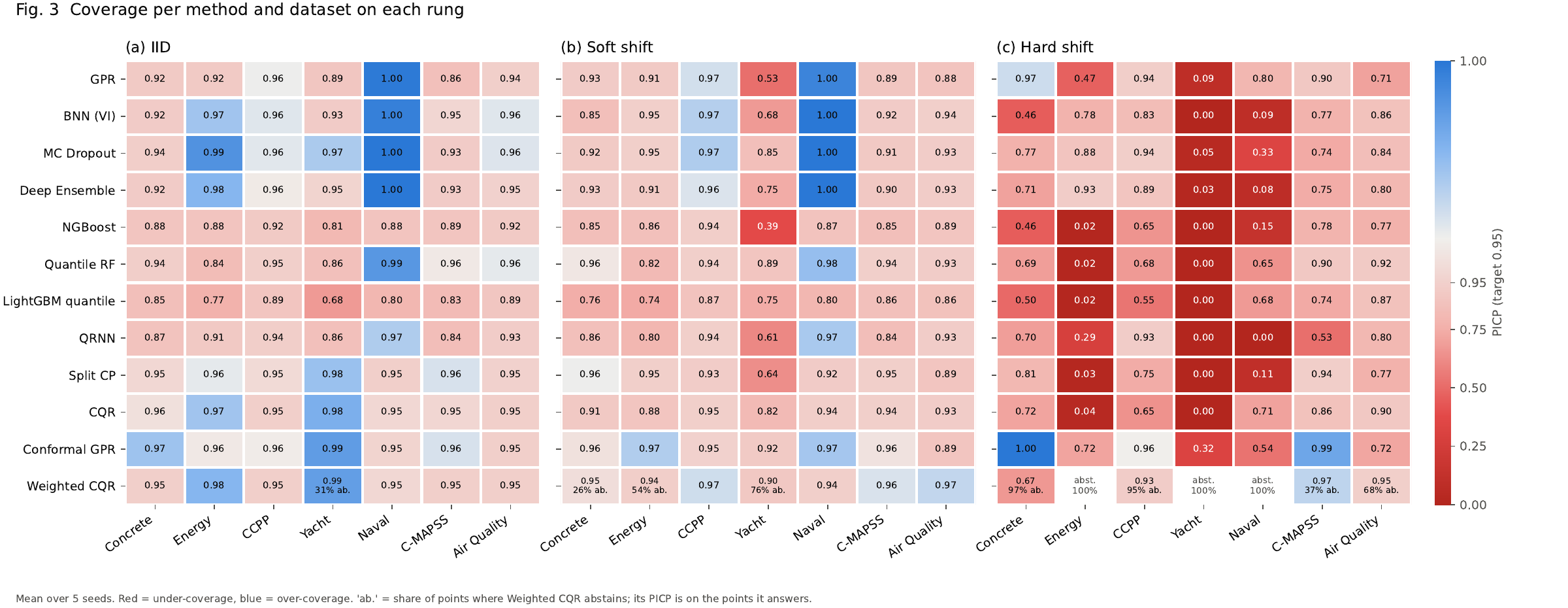}
\caption{Coverage per method and dataset: (a) IID, (b) soft shift, (c) hard shift. Mean over 5 seeds; red = under-coverage, blue = over-coverage. `ab.': Weighted CQR abstention rate; its PICP is on answered points.}
\label{fig:heatmaps}
\end{figure}

On the IID rung (Fig.~\ref{fig:heatmaps}a), all four conformal methods reach near-nominal coverage on every dataset: the lowest dataset-level PICP is 0.946 (Weighted CQR on C-MAPSS), and Split CP, CQR and Conformal GPR never fall below 0.947. This is the exchangeable setting their guarantee covers, and it is the setting random-split benchmarks measure.

The neural Bayesian and ensemble methods come close without a guarantee. BNN, MC Dropout and the deep ensemble average 0.96 and stay at or above 0.92 on all seven datasets, their lowest values on Concrete or C-MAPSS. The GP averages 0.93 but falls to 0.86 on C-MAPSS. The tree-based and distributional methods under-cover systematically. The LightGBM quantile model averages 0.81 and reaches 0.90 on no dataset (0.68 on Yacht). NGBoost averages 0.88 and QRNN 0.90, both lowest on the small Yacht set or on C-MAPSS. Quantile RF averages 0.93 but drops to 0.84 on Energy.

These differences are real but small beside what follows. Interval quality is not distinguishable across methods on this rung: the Friedman test on Winkler scores gives $p = 0.23$.

\subsection{Under soft shift, average coverage erodes modestly but subgroup coverage fails}\label{sec:res-soft}

When source and target overlap, the average loss is small (Fig.~\ref{fig:heatmaps}b). Against each method's own IID test set, only three drops survive Holm correction: NGBoost (0.88 to 0.81), Split CP (0.95 to 0.89) and CQR (0.95 to 0.91). Six methods show a drop at the unadjusted 5\% level. The two tree quantile methods show none, but they were already under-covering in distribution.

\begin{figure}[pos=tbp]
\centering
\includegraphics[width=0.85\linewidth,trim=0 36 0 19,clip]{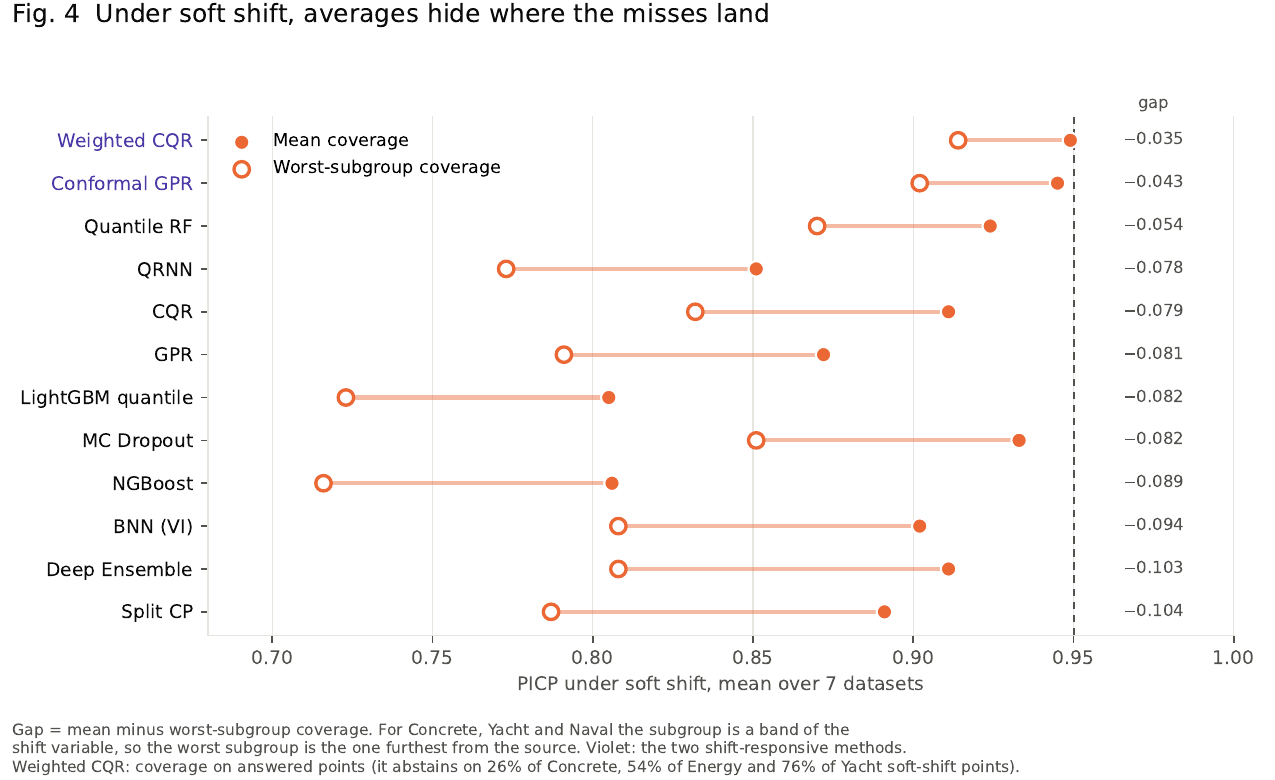}
\caption{Gap between mean and worst-subgroup coverage under soft shift. For Concrete, Yacht and Naval the subgroups are bands of the shift variable. Violet: the two shift-responsive methods. Weighted CQR: coverage on answered points.}
\label{fig:subgroup}
\end{figure}

The average hides where the misses land. Apart from the two methods discussed below, every method loses 0.05--0.10 of coverage between its mean and its worst subgroup (Table~\ref{tab:summary}, Fig.~\ref{fig:subgroup}). The deep ensemble covers 0.91 on average but 0.81 in its worst subgroup. The GP covers 0.87 on average and 0.79 in its worst subgroup, and Split CP 0.89 and 0.79. For Concrete, Yacht and Naval the subgroup variable is the shift variable, so the worst subgroup is the part of the target furthest from the source, which is where an engineer most needs the interval. Only the two methods whose intervals respond to the shift keep the gap small: Conformal GPR (0.94 on average, 0.90 worst-group), whose width scales with the GP's posterior variance, and Weighted CQR (0.95 and 0.91), which reweights its calibration set.

On this rung too, the Friedman test detects no difference in Winkler scores across methods ($p = 0.24$). NGBoost (4.3) and Quantile RF (4.4) have the best mean ranks, NGBoost despite 0.81 coverage because its intervals are narrow; the ordering is descriptive only, and it shows why Winkler rank alone should not be read as a verdict on reliability.

\subsection{Under hard shift, coverage collapses and rankings dissolve}\label{sec:res-hard}

On the hard rung, mean coverage across datasets falls to between 0.40 (NGBoost) and 0.75 (Conformal GPR) for the 11 methods that always answer (Fig.~\ref{fig:coverage-ladder}, Fig.~\ref{fig:heatmaps}c). Nine of the 11 lose coverage significantly relative to their own IID test set across seeds after Holm correction, with mean drops of 0.32--0.48. The two exceptions are GPR (drop 0.23, Holm-adjusted $p = 0.20$) and Conformal GPR (0.21, $p = 0.11$). Their drop is not consistent, not small: they over-cover on some datasets (Conformal GPR reaches 1.00 on Concrete and 0.99 on C-MAPSS) and collapse on others (0.32 on Yacht). Section~\ref{sec:res-mech} explains why.

The type of hard shift matters more than the method. On the five extrapolation datasets, mean coverage across the 11 methods is 0.46, ranging from 0.00 to 1.00. On the two conditional shifts it is 0.81 (C-MAPSS and Air Quality, 0.53--0.99). Every method is further from nominal under extrapolation than under conditional shift. Air Quality is the mildest hard shift in the benchmark (0.71--0.92), and there the GP methods are the worst performers (0.71--0.72), the reverse of their standing on the extrapolation datasets.

\begin{figure}[pos=tbp]
\centering
\includegraphics[width=\linewidth,trim=0 19 0 19,clip]{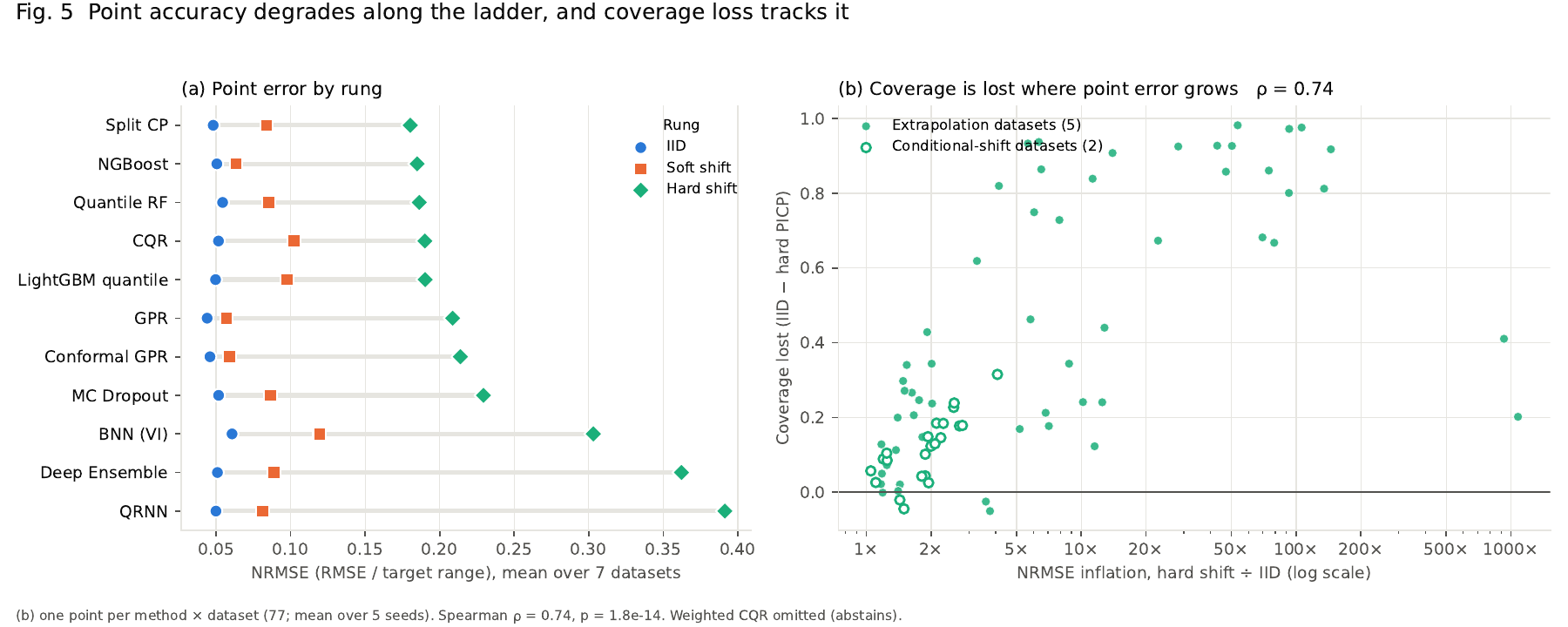}
\caption{(a) NRMSE by rung. (b) IID-to-hard coverage drop against NRMSE inflation; one point per method and dataset (77, mean over 5 seeds). Weighted CQR omitted.}
\label{fig:nrmse}
\end{figure}

Intervals fail where point predictions fail (Fig.~\ref{fig:nrmse}). Mean NRMSE is 0.044--0.061 in distribution for every method and rises to 0.18--0.39 under hard shift. How much depends on the dataset: the median NRMSE of the 11 non-abstaining methods grows 1.4-fold on CCPP and C-MAPSS, 2-fold on Air Quality, 3.6-fold on Concrete, 5.6-fold on Energy, 14-fold on Naval and 70-fold on Yacht. Across the 77 method--dataset pairs, the growth in NRMSE predicts the coverage lost (Spearman $\rho = 0.74$ between log NRMSE inflation and the IID-to-hard PICP drop, $p < 10^{-13}$). Point accuracy does not rank methods the way coverage does. The three methods with the lowest hard-shift NRMSE (Split CP 0.180, NGBoost 0.185, Quantile RF 0.186) cover only 0.40--0.55, because their intervals do not widen as their errors grow. The deep ensemble and QRNN have the worst point accuracy (0.36 and 0.39).

\begin{figure}[pos=tbp]
\centering
\includegraphics[width=\linewidth,trim=0 28 0 19,clip]{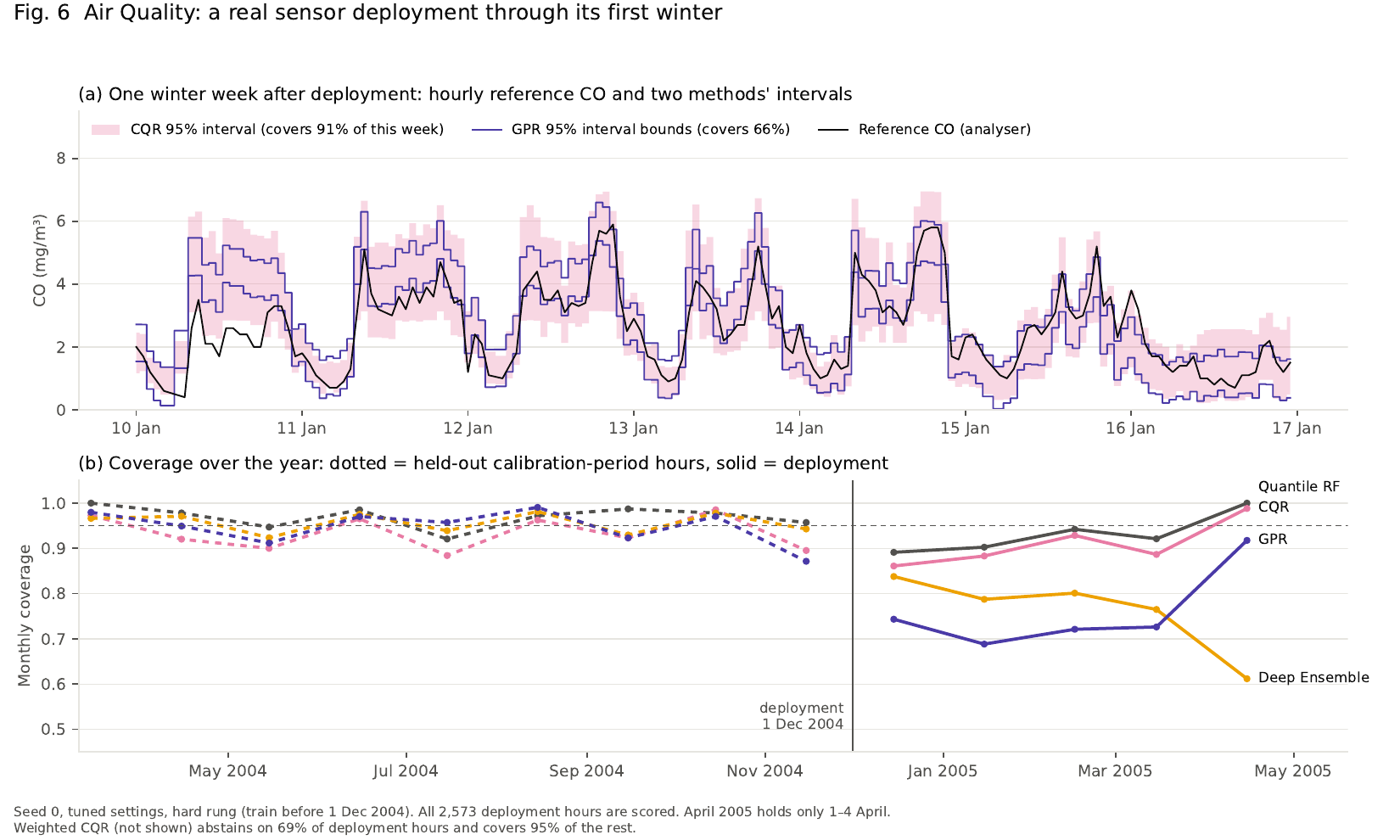}
\caption{Air Quality deployment (seed~0, tuned settings, hard rung; training before 1 December 2004). (a) Hourly reference CO and two methods' intervals over one winter week. (b) Monthly coverage: dotted = held-out calibration-period hours, solid = deployment hours. April 2005 covers 1--4 April only.}
\label{fig:airquality}
\end{figure}

On Air Quality the failure is persistent, not gradual (Fig.~\ref{fig:airquality}). Refitting the tuned models for seed~0 and scoring every one of the 2,573 deployment hours, monthly coverage from December to March stays at 0.69--0.74 for the GP and 0.86--0.93 for CQR, against 0.93--0.97 for all four methods on held-out hours from the calibration period. The deep ensemble stays at 0.77--0.84 from December to March (the April bin holds only four days). The shift arrives with the season, so a model validated on the months before December gives no warning of it.

No method has the lowest Winkler score on more than two datasets, and every method's rank moves across datasets (Fig.~S3). The Friedman test detects no systematic difference in ranks ($p = 0.63$). The lowest Winkler score belongs to a tree-based or conformalised-tree method on four datasets (Split CP on Concrete, CQR on Naval, Quantile RF on C-MAPSS and Air Quality), a neural method on two (the deep ensemble on Energy, QRNN on CCPP), and Conformal GPR on Yacht, where its coverage is only 0.32 and no method is usable. Tuning changes the winner on two of the seven datasets (Section~\ref{sec:res-tuning}): with fixed settings, MC Dropout had the lowest Winkler score on Energy and on Air Quality.

\subsection{Three mechanisms explain the failures}\label{sec:res-mech}

\begin{figure}[pos=tbp]
\centering
\includegraphics[width=\linewidth,trim=0 17 0 19,clip]{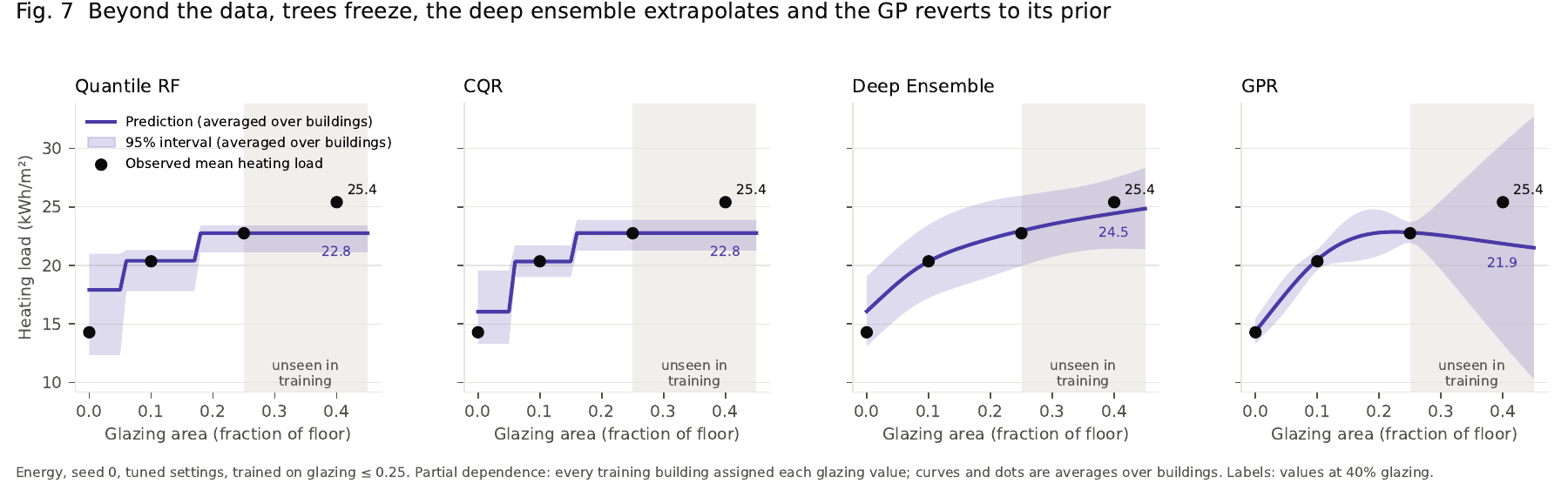}
\caption{Hard-shift predictions on Energy (seed~0, tuned settings, trained on glazing $\le 0.25$). Partial dependence: each training building is assigned every glazing value, and curves and dots average over buildings. Labels give values at 40\% glazing.}
\label{fig:energy}
\end{figure}

\textbf{Tree-based intervals extrapolate as constants.} A tree ensemble predicts the value of its outermost leaf for any input beyond its last split, and its quantile band is copied from the same leaves. When the target lies outside the training range, both the point prediction and the band freeze while the truth moves on. Energy shows this cleanly. Heating load rises with glazing area, the target domain is the one glazing level above the training range, and coverage there is 0.02 for Quantile RF, the LightGBM quantile model and NGBoost, 0.03 for Split CP and 0.04 for CQR. Conformal calibration cannot help, because it shifts a band that sits in the wrong place by an amount learned where the band was right. On Yacht, where resistance rises steeply beyond Froude 0.35, every tree-based method covers 0.00. Fig.~\ref{fig:energy} shows the mechanism directly. Averaged over buildings, Quantile RF and CQR predict 22.8\,kWh/m$^2$ at 40\% glazing, the value they learned at 25\%, against an observed 25.4. The deep ensemble extrapolates the trend (24.5), and the GP turns back towards its prior mean (21.9) while its interval widens.

\begin{figure}[pos=t]
\centering
\includegraphics[width=\linewidth,trim=0 17 0 19,clip]{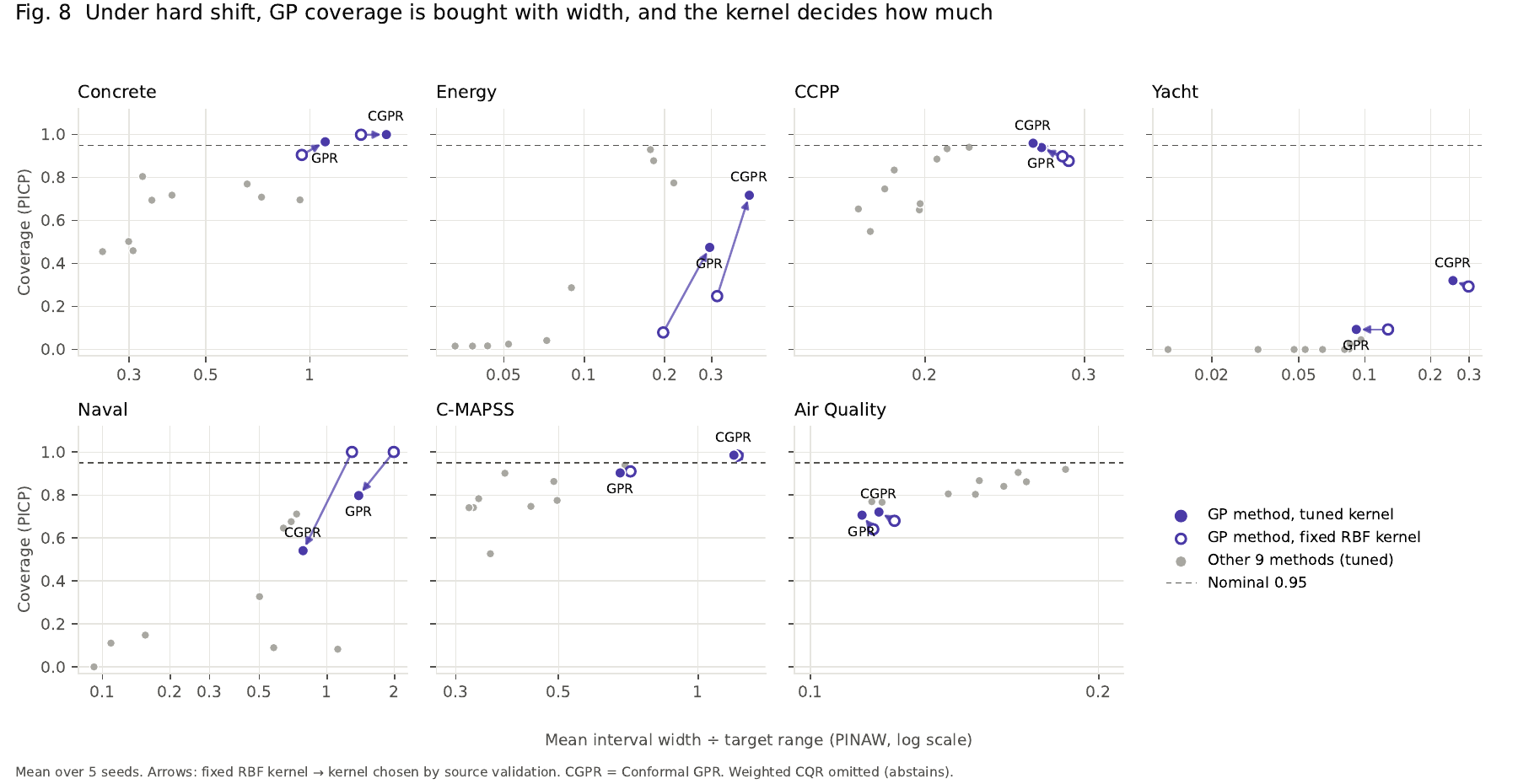}
\caption{GP interval width and coverage under hard shift (mean over 5 seeds). Arrows: fixed RBF kernel $\to$ kernel chosen by source validation. CGPR: Conformal GPR.}
\label{fig:gpkernel}
\end{figure}

\textbf{GP coverage is bought with width, and the kernel decides how much.} A stationary GP reverts to its prior away from the data, so its variance grows with distance and its intervals widen. That is why the GP methods have the highest hard-shift coverage in Table~\ref{tab:summary}, and also the widest intervals (mean PINAW 0.56 and 0.67, against 0.15--0.41 for the rest). Fig.~\ref{fig:gpkernel} shows the trade-off per dataset, and shows that it is set by the kernel. With the fixed RBF kernel, the GP covered every Naval target point, with intervals twice the target range (PINAW 1.99), which is a correct but useless statement. Tuning chose Mat\'ern kernels on Naval in all five runs; coverage fell to 0.80 at PINAW 1.39, and Conformal GPR fell from 1.00 to 0.54. On Energy the same choice moved coverage the other way, from 0.08 to 0.48 for GPR and from 0.25 to 0.72 for Conformal GPR. How fast the kernel's correlation decays governs behaviour beyond the data, and source data carry no information about that region, so the kernel that wins on source validation is not the one that behaves well under extrapolation (Section~\ref{sec:res-tuning}).

\begin{figure}[pos=tbp]
\centering
\includegraphics[width=0.8\linewidth,trim=0 30 0 19,clip]{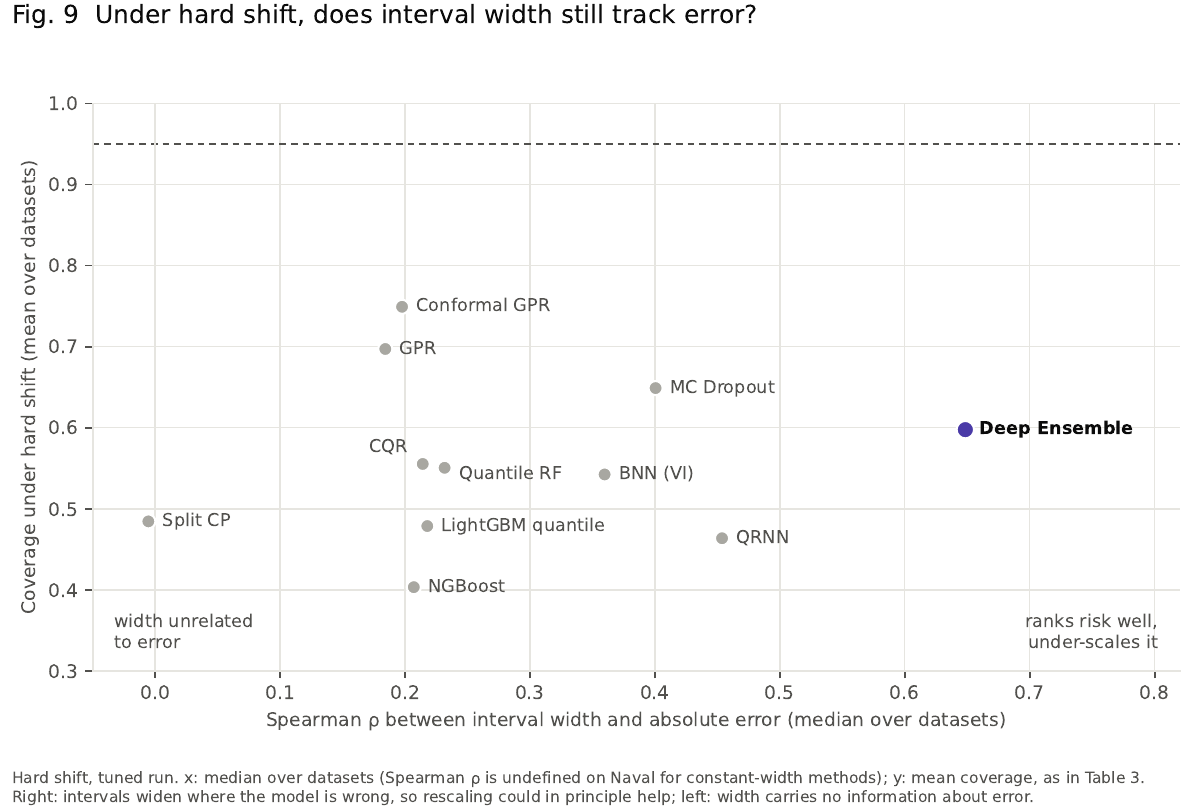}
\caption{Width--error Spearman correlation against coverage under hard shift (tuned run). $x$: median over datasets ($\rho$ is undefined on Naval for constant-width methods); $y$: mean coverage, as in Table~\ref{tab:summary}.}
\label{fig:widtherror}
\end{figure}

\textbf{Ensembles rank risk correctly but under-scale it.} Under hard shift the deep ensemble's interval width still tracks its error: the Spearman correlation between width and absolute error has a median of 0.65 across datasets (mean 0.52; Fig.~\ref{fig:widtherror}). Yet its coverage is 0.60. Members disagree more where they are wrong, but not by enough. A rescaling fitted on target residuals could therefore recover coverage without losing this ordering; whether one fitted on source calibration data would transfer to the target is not tested here. The tree failure is different: Split CP's constant width cannot track error at all (median $\rho\approx0$), and the tree quantile methods track it weakly (median $\rho$ 0.22--0.23), so no rescaling could make their width informative.

\subsection{Weighted conformal holds coverage by abstaining, and its abstention is driven by sample size, not weight error}\label{sec:res-wcqr}

Weighted CQR is the only method whose coverage on the points it answers stays near nominal on every rung, with one exception (Concrete under hard shift), and Table~\ref{tab:wcqr} shows the price. Where the effective calibration sample is large, it answers almost everywhere and covers 0.94--0.97. Where the sample is small, it declines to answer.

\begin{table*}[width=\textwidth,cols=10,pos=t]
\centering
\caption{Weighted CQR per dataset and rung (tuned run, mean over 5 seeds). ESS is the effective calibration sample size; PICP is on the points it answers; ``--'' means it abstains on every point.}
\label{tab:wcqr}
\footnotesize
\setlength{\tabcolsep}{4pt}
\begin{tabular}{@{}lccccccccc@{}}
\toprule
 & \multicolumn{3}{c}{IID} & \multicolumn{3}{c}{Soft shift} & \multicolumn{3}{c}{Hard shift} \\
\cmidrule(lr){2-4}\cmidrule(lr){5-7}\cmidrule(l){8-10}
Dataset & ESS & Abstain & PICP & ESS & Abstain & PICP & ESS & Abstain & PICP \\
\midrule
Concrete & 73 & 1\% & 0.95 & 38 & 26\% & 0.95 & 60 & 97\% & 0.67 \\
Energy & 64 & 0\% & 0.98 & 12 & 54\% & 0.94 & 47 & 100\% & -- \\
CCPP & 593 & 0\% & 0.95 & 172 & 0\% & 0.97 & 46 & 95\% & 0.93 \\
Yacht & 19 & 31\% & 0.99 & 7 & 76\% & 0.90 & 10 & 100\% & -- \\
Naval & 594 & 0\% & 0.95 & 130 & 0\% & 0.94 & 86 & 100\% & -- \\
C-MAPSS & 344 & 0\% & 0.95 & 203 & 1\% & 0.96 & 367 & 37\% & 0.97 \\
Air Quality & 579 & 0\% & 0.95 & 125 & 0\% & 0.97 & 33 & 68\% & 0.95 \\
\bottomrule
\end{tabular}
\end{table*}

Under hard shift it abstains on 85\% of target points on average over the seven datasets (macro-average; 77\% when pooled over all evaluated target points), and on 98\% on average across the five extrapolation datasets. On those five, abstention is the appropriate response: the density ratio is unbounded, the calibration set says nothing about the target region, and no finite interval carries a guarantee. On the two conditional shifts the covariates overlap, so the classifier assigns finite weights. It abstains on 37\% of C-MAPSS points and 68\% of Air Quality points and covers 0.97 and 0.95 of the rest. Answering can still be wrong. On Concrete it answers only 3\% of target points (about 8 per seed) and covers 0.67 of those; on CCPP it answers 5\% (about 100 per seed) and covers 0.93.

\begin{figure}[pos=tbp]
\centering
\includegraphics[width=\linewidth,trim=0 19 0 19,clip]{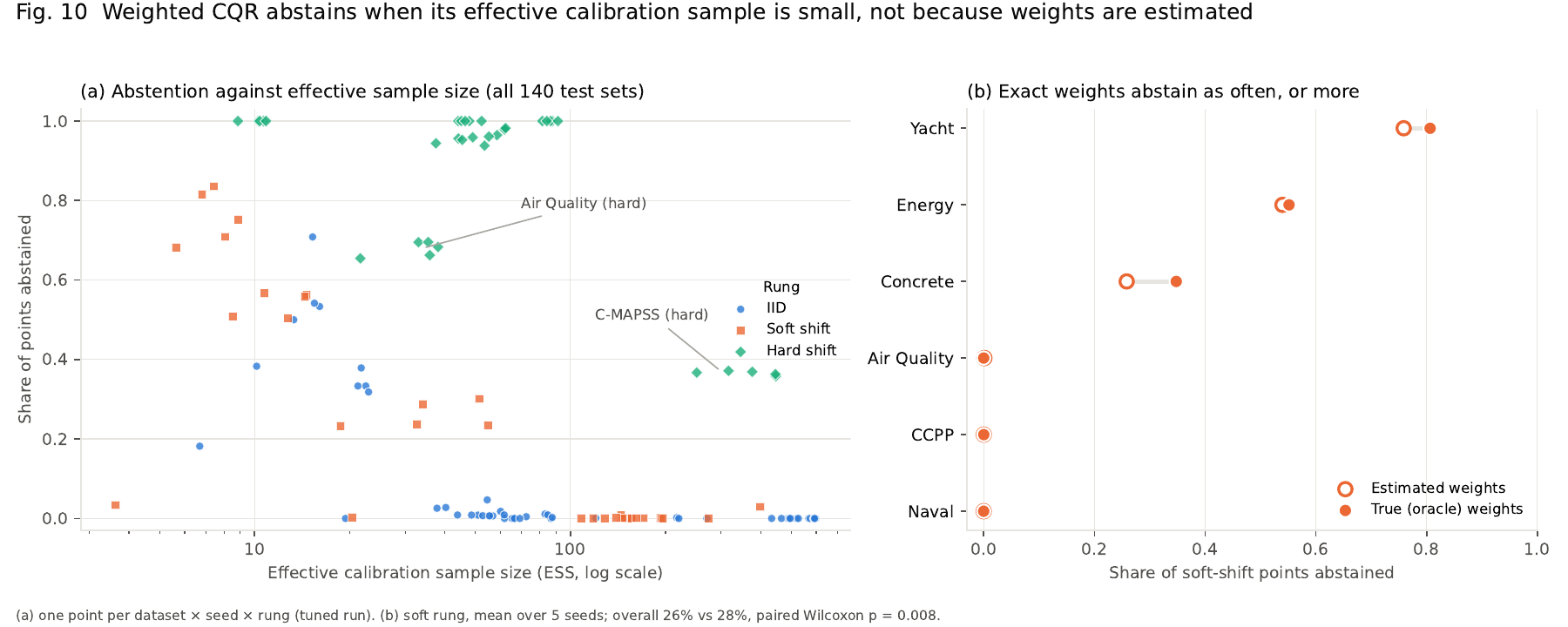}
\caption{(a) Weighted CQR abstention against effective sample size; one point per dataset, seed and rung (140 test sets, tuned run). (b) Abstention with exact versus estimated weights on the soft rung, mean over 5 seeds.}
\label{fig:wcqr}
\end{figure}

Under soft shift, abstention follows the effective sample size (Fig.~\ref{fig:wcqr}a). It abstains on 54\% of Energy points (ESS 12), 76\% of Yacht points (ESS 7) and 26\% of Concrete points (ESS 38), and on at most 1\% of points in the four datasets with ESS above 120. Small calibration sets cause it even without shift: on Yacht's IID rung (ESS 19) it abstains on 31\% of points. What abstention buys is visible against unweighted CQR on the same fitted models: on the six datasets with a logistic soft rung, CQR covers 0.91 on average and 0.83 in its worst subgroup, and Weighted CQR 0.95 and 0.92.

The oracle-weight ablation separates the cost of estimating weights from the cost of small samples. On those six datasets the true density ratio is known in closed form, $w\propto\exp(1.5\,z(v))$, so the same fitted CQR can be calibrated with exact weights. With exact weights (Fig.~\ref{fig:wcqr}b), the method abstains \emph{more}, not less: 28\% of points against 26\% with estimated weights (paired Wilcoxon over 30 dataset--seed pairs, $p = 0.008$). Coverage on answered points is statistically unchanged (0.94 against 0.95, $p = 0.24$). Most of the difference is on Concrete (35\% against 26\%), where the exact weights give a smaller effective sample (ESS 28 against 38). Abstention under soft shift is therefore a small-ESS phenomenon of weighted conformal prediction itself, not an artefact of density-ratio estimation. On Air Quality, where selection acts on time and the estimated weights see only $x$, the estimated-weight version over-covers (0.97 against 0.95 with oracle weights) with wider intervals (PINAW 0.22 against 0.18): conservative, not unsafe.

\subsection{Calibrating on rows instead of whole engines breaks conformal coverage even without shift}\label{sec:res-grouped}

On C-MAPSS we refitted the conformal methods with the calibration set drawn at random from all training rows, so each engine's cycles are split between proper training and calibration, and compared them with the benchmark's engine-level calibration. Splits, seeds and tuned settings are identical, and the engine-level runs reproduce the benchmark exactly.

\begin{table*}[width=\textwidth,cols=8,pos=t]
\centering
\caption{C-MAPSS, IID rung: conformal methods with engine-level versus row-level calibration (5 seeds $\times$ 2 ladder modes). $\PICP_{\min}$: worst remaining-useful-life band. $p$: paired Wilcoxon test on PICP. Fig.~S2 shows all three rungs.}
\label{tab:cmapss}
\footnotesize
\setlength{\tabcolsep}{4pt}
\begin{tabular}{@{}lccccccc@{}}
\toprule
 & \multicolumn{2}{c}{PICP} & \multicolumn{2}{c}{$\PICP_{\min}$} & \multicolumn{2}{c}{PINAW} & \\
\cmidrule(lr){2-3}\cmidrule(lr){4-5}\cmidrule(lr){6-7}
Method & Engine & Row & Engine & Row & Engine & Row & $p$ \\
\midrule
Split CP & 0.96 & 0.90 & 0.90 & 0.78 & 0.66 & 0.50 & 0.002 \\
CQR & 0.95 & 0.93 & 0.85 & 0.81 & 0.54 & 0.48 & 0.016 \\
Conformal GPR & 0.96 & 0.91 & 0.91 & 0.80 & 0.74 & 0.52 & 0.004 \\
Weighted CQR & 0.95 & 0.94 & 0.86 & 0.81 & 0.54 & 0.49 & 0.027 \\
\bottomrule
\end{tabular}
\end{table*}

Row-level calibration loses 2--6 points of coverage on unseen engines and up to 12 points in the worst RUL band, while making the intervals 10--30\% narrower. Rows from an engine the model has trained on are easier to predict than rows from a new engine, so the calibration residuals are too small and the conformal quantile is too tight. The guarantee is intact in form but void in substance, because calibration and test points are no longer exchangeable. The same direction holds under both shift rungs (for example, Split CP under hard shift: 0.94 to 0.88), though with five runs per rung the differences do not reach significance ($p \ge 0.06$). For any engineering dataset with repeated measurements of the same asset, calibration must be split by asset.

\subsection{Source-domain tuning improves IID and soft-shift intervals but gives no consistent benefit under hard shift}\label{sec:res-tuning}

Nested tuning moved almost every method off its defaults. In the 35 hard-mode runs, the default configuration was kept in at most 20\% of runs for every searched method except Quantile RF (49\%) and the point LightGBM behind Split CP (26\%). The GP chose a Mat\'ern-3/2 kernel in 20 runs, Mat\'ern-5/2 in 8 and the default RBF in 7.

In distribution, tuning helps. After Holm correction across the 36 method $\times$ rung tests, Winkler scores improve significantly for the LightGBM quantile model (median change $-14\%$), NGBoost ($-7\%$), CQR ($-7\%$), Weighted CQR ($-5.5\%$) and BNN ($-3\%$). Under soft shift, one improvement survives correction, NGBoost ($-15\%$), and five more improve at the unadjusted level: LightGBM quantile ($-11\%$), Weighted CQR ($-7\%$), CQR ($-6\%$), BNN ($-3\%$) and the deep ensemble ($-1\%$). Under hard shift the median change is 0\% for most methods. Only NGBoost improves after correction ($-5\%$), and CQR gets slightly worse ($+2\%$, unadjusted $p = 0.04$).

The reason is that what tuning learns on source data does not transfer to the hard rung (Fig.~S1). Across the 8 searched methods $\times$ 7 datasets $\times$ 5 seeds ($n = 280$ per rung), the improvement found on inner source validation predicts the improvement on the IID test set (Spearman $\rho = 0.33$, $p < 10^{-7}$) and on the soft-shift test set ($\rho = 0.40$, $p < 10^{-11}$), but not on the hard-shift test set ($\rho = 0.04$, $p = 0.48$). The tuned model beats the default in 59\% and 58\% of cases on the first two rungs, and in 49\% under hard shift. Behind this average of zero are large swings of both signs. The same kernel choice made Conformal GPR's Winkler score 419\% worse on Naval and 60\% better on Energy (Section~\ref{sec:res-mech}). This is the model-selection counterpart of Section~\ref{sec:theory}: under extrapolation, source data carry no information about the region that matters, so neither calibration nor validation on them can protect the target.

The benchmark's conclusions do not depend on tuning. With fixed settings, mean hard-shift coverage spans 0.37--0.73 (tuned: 0.40--0.75), Weighted CQR abstains on 85\% of hard-shift points on average (tuned: 85\%), and no rung gives a significant Friedman result (hard shift: $p = 0.49$ fixed, 0.63 tuned). The per-method comparison is in Table~S1 of the supplement.

\section{Discussion}\label{sec:discussion}

The central result is that prediction intervals calibrated on a random split did not survive extrapolation for any of the evaluated non-selective methods, and source-domain tuning did not make them do so. Under hard shift, coverage falls to 0.40--0.75 for every method that always answers, tuning on source data changes the outcome unpredictably, and the only method that stays near nominal on the points it answers does so by declining to predict on most of them. This section turns those results into guidance for practitioners (Section~\ref{sec:recs}), separates what is impossible from what is merely unsolved (Sections~\ref{sec:impossible} and~\ref{sec:open}), and states the limits of the evidence (Section~\ref{sec:limits}).

\subsection{Practical recommendations}\label{sec:recs}

Six recommendations follow directly from Section~\ref{sec:results}. Each is stated with the result that supports it.

\begin{enumerate}
\item \textbf{Evaluate UQ on a split along the deployment variable, not only on a random split.} On random splits the conformal methods cover at least 0.946 on every dataset; along the shift ladder the same methods fall to 0.48--0.75 on average under hard shift. A random-split benchmark measures the one setting in which conformal guarantees hold by construction. For an engineering model, the split should follow the variable along which the model will be used: age, design parameter, operating condition, fault mode or calendar time.
\item \textbf{Report sharpness, subgroup coverage and point error next to marginal coverage.} Marginal coverage hid a 0.05--0.10 shortfall in the worst subgroup under soft shift (Fig.~\ref{fig:subgroup}), and full coverage on Naval came from intervals twice the target range (Fig.~\ref{fig:gpkernel}). Point error is also an inexpensive warning signal: growth in NRMSE tracked coverage loss closely ($\rho = 0.74$, Fig.~\ref{fig:nrmse}). Once a few labelled observations arrive after deployment, a rise in residuals relative to the validation level is evidence that the intervals have stopped being reliable.
\item \textbf{Preserve the deployment unit when splitting calibration data.} When the data contain repeated measurements of the same engine, structure, sensor or specimen, split training, calibration and test data by that unit, not by row. Row-level calibration on C-MAPSS cost 2--6 points of coverage on new engines and up to 12 points near end of life, while producing narrower and more convincing intervals (Table~\ref{tab:cmapss}).
\item \textbf{Do not rely on tree-based intervals, conformalised or not, beyond the training range of an input.} Trees predict a constant beyond their last split, so both the point prediction and the band freeze. Coverage was 0.02--0.04 on Energy and 0.00 on Yacht for every tree-based method (Fig.~\ref{fig:energy}). Conformal calibration does not repair this, because the residual quantile is learned where the band was right.
\item \textbf{Treat abstention as a legitimate safety response, report its rate and monitor the effective sample size.} Weighted CQR abstained on 98\% of points on the extrapolation datasets, an appropriate response when the calibration data carry no information about the target. Under soft shift, its abstention rose sharply once the effective calibration sample fell to a few tens (Fig.~\ref{fig:wcqr}a), so the ESS should be reported with every weighted conformal deployment. A downstream decision process needs an explicit route for abstained cases: a conservative default, an engineering margin, human review, or a request for a measurement. Abstention transfers the decision rather than resolving the uncertainty, so its frequency is an operational cost to be reported alongside coverage.
\item \textbf{Do not expect validation on source data to select settings that are robust under shift.} Tuning gains carried over to the IID and soft-shift test sets but not to hard shift ($\rho = 0.04$, Fig.~S1), and the GP kernel chosen on source data moved hard-shift coverage by up to 0.47 in either direction. Where extrapolation is expected, report results for several kernels or architectures rather than a single tuned configuration.
\end{enumerate}

\subsection{What source data cannot guarantee}\label{sec:impossible}

Under extrapolation, the failure is not a weakness of particular methods but a property of the problem. When the target distribution puts mass where the source has none, the density ratio is unbounded and the calibration residuals say nothing about the residuals that matter. No procedure that learns only from source data can then guarantee coverage. What each method does instead is decided by its inductive bias off the support (Section~\ref{sec:theory}, Fig.~\ref{fig:energy}). The GP reverts to its prior and covers by being wide; the tree methods freeze and miss; the networks extrapolate a trend that may or may not hold. Which bias happens to match the physics varies by dataset, so the rankings dissolve (Fig.~S3) and no family is safe. Weighted conformal prediction is the only evaluated method designed to respond to this: it abstains.

The tuning result is the same fact seen from model selection. Choosing a kernel or an architecture is a decision about behaviour beyond the data, and validation rows drawn from the source cannot inform it. That is why tuning gains carry over to the soft rung, where the source still covers the target region, and vanish under hard shift. Importance-weighted validation \citep{sugiyama2007covariate} does not escape this, because it relies on the same density ratio that is unbounded here.

Conditional shift is different and in one respect worse. On C-MAPSS and Air Quality the covariates overlap, so the density ratio is finite and Weighted CQR answers part of the target: 63\% of C-MAPSS points (ESS 367) and 32\% of Air Quality points (ESS 33). It covered 0.97 and 0.95 of the points it answered, but nothing guarantees this: reweighting $X$ cannot correct a change in $P(Y\mid X)$, and the coverage gap bounds of \citet{barber2023conformal} grow with exactly that change. A covariate-based check, whether a classifier, a density ratio or the ESS, cannot detect a shift that leaves the covariates unchanged. The Concrete result shows the converse risk: even when Weighted CQR abstains on 97\% of points, the 3\% it answers can be wrong (coverage 0.67).

The practical consequence is that guarantees beyond the data must come from outside the data. Three sources are available to an engineer: physical constraints on the response (monotonicity in age or speed, known asymptotes, conservation laws); a small number of labelled measurements from the target regime; or an explicit decision to abstain and defer to engineering judgement. A UQ method can only report which of these the situation calls for.

\subsection{Open problems}\label{sec:open}

Three problems are unsolved rather than impossible, and the benchmark locates each one.

\textbf{Coverage with a small effective sample.} Under soft shift the density ratio exists and weighted conformal prediction is valid in principle, yet it abstained on up to 76\% of points. The oracle-weight ablation shows that this is not caused by estimating the weights: exact weights abstained slightly more (Fig.~\ref{fig:wcqr}b). The constraint is the effective sample size, which reweighting shrinks. Useful progress would therefore come from methods whose coverage degrades gracefully as the ESS falls rather than switching to abstention. One route is to trade a controlled loss of coverage for fewer abstentions, in the spirit of conformal risk control \citep{angelopoulos2024conformal}. Another is to bring in structural assumptions, such as smoothness of the residual distribution along the shift variable, and state the guarantee conditional on them. Our ongoing work develops guarantees of this kind under estimated weights; the present benchmark supplies the baseline any such method must beat.

\textbf{Detecting conditional shift.} C-MAPSS and Air Quality show that the most dangerous shift for a deployed model is the one its covariates do not reveal. Detecting it requires labels. Air Quality coverage fell in the first weeks of deployment and stayed low (Fig.~\ref{fig:airquality}), which is the setting adaptive conformal methods were designed for \citep{gibbs2021adaptive}: they update the interval online as labelled outcomes arrive. How few labels such methods need to restore coverage after a regime change in engineering data is an open, testable question, and Air Quality is a natural benchmark for it.

\textbf{Model selection under shift.} Source validation chose GP kernels that roughly halved Conformal GPR's coverage on Naval (1.00 to 0.54) and nearly tripled it on Energy (0.25 to 0.72). A selection criterion that accounts for behaviour beyond the data, for example by preferring kernels and architectures whose extrapolation agrees with known physical trends, would turn this from luck into design. The width--error result points to a related, cheaper question. The deep ensemble ranks its errors correctly under hard shift but under-scales them (Fig.~\ref{fig:widtherror}), so post-hoc recalibration \citep{kuleshov2018accurate} might recover part of the lost coverage; whether a rescaling fitted on source data transfers to the target is untested here.

\subsection{Limitations}\label{sec:limits}

The evidence has the following limits, roughly in order of how much they could change the conclusions.

\begin{itemize}
\item \textbf{Scope of shift.} Each dataset has a single shift variable and a single threshold, and the soft rung uses one steepness ($k = 1.5$). A sweep over $k$ and over thresholds would turn the ladder into a severity curve. The conclusions about hard shift rest on the ladder's endpoints, which do not depend on $k$.
\item \textbf{Data provenance.} Three datasets come from simulators (Energy, Naval, C-MAPSS) and one from towing-tank tests (Yacht). Concrete, CCPP and Air Quality are physical measurements, and only Air Quality is a field deployment over time.
\item \textbf{Number of datasets and seeds.} Seven datasets give dataset-level tests little power. A non-significant Friedman result is therefore read as no detectable difference, not as equivalence. The coverage-drop tests pair 35 dataset--seed runs; five seeds capture split and initialisation variability within each dataset but not variability across deployment environments, so those tests show that drops are consistent across seeds on these datasets, not that they generalise to other datasets. The per-dataset results (Fig.~\ref{fig:heatmaps}, Fig.~S3) carry the argument that rankings are unstable.
\item \textbf{Heterogeneous hard shift.} The hard rung groups support extrapolation, an unseen fault mode and a temporal deployment. These share operational severity but not a formal shift mechanism, which is why the extrapolation and conditional datasets are reported separately where the distinction matters.
\item \textbf{No recalibration baseline.} We did not evaluate post-hoc variance scaling or other recalibration of the non-conformal probabilistic methods. The experiments therefore do not show whether a rescaling fitted on source data would recover hard-shift coverage for methods, such as the deep ensemble, whose interval width still tracks error.
\item \textbf{Selective coverage.} Weighted CQR's coverage on answered points is conditional on answering and is not unconditional deployment coverage. Under hard shift it rests on few points on some datasets (about 8 per seed on Concrete).
\item \textbf{Model scale and tuning budget.} Training sets are capped at 3,000 rows and test sets at 2,000, tuning fits at 1,500 rows, and the search uses 8 random candidates per method (3 kernels for the GP). Larger budgets could change individual results, but the tuned-versus-fixed comparison (Section~\ref{sec:res-tuning}) suggests the qualitative conclusions are not sensitive to them.
\item \textbf{Model choices.} C-MAPSS is modelled with smoothed tabular features rather than a sequence model. Weighted CQR uses one density-ratio estimator (degree-2 logistic regression) and is transductive, since it sees the unlabelled test batch. The tuning criterion is the Winkler score, and a likelihood-based criterion could select different configurations for the Gaussian methods.
\item \textbf{Targets and levels.} All targets are univariate and all intervals are at 95\%. Behaviour at other levels, and for joint prediction of several responses, is untested.
\item \textbf{Illustrations.} Figs.~\ref{fig:airquality} and~\ref{fig:energy} come from refits for one seed. The quantities they illustrate (Air Quality coverage, tree extrapolation on Energy) agree with the five-seed results in Fig.~\ref{fig:heatmaps}.
\end{itemize}

\section{Conclusion}\label{sec:conclusion}

We evaluated 12 uncertainty quantification methods from five families on seven engineering regression datasets along a three-rung shift ladder: in distribution, soft covariate shift, and hard shift by extrapolation or by a change of regime. Every method was tuned on source data only.

Calibration guarantees do not survive extrapolation. In distribution, the conformal methods covered at least 0.946 on every dataset. Under soft shift, average coverage held up but the worst subgroups lost 0.05--0.10. Under hard shift, mean coverage fell to 0.40--0.75 for every method that always answers, and no systematic difference in method ranks was detected. Three mechanisms explain the failures. Tree-based intervals freeze beyond the training range; GP intervals cover by growing wide, at a cost set by a kernel that source data cannot choose; and ensembles rank their errors correctly but under-scale them. Coverage loss tracked the growth in point error ($\rho = 0.74$). Tuning improved interval quality in distribution and under soft shift, but its gains did not carry over to hard shift ($\rho = 0.04$).

Weighted conformal prediction was the only method that recognised extrapolation: it abstained on 98\% of points on the extrapolation datasets. With exact density-ratio weights it abstained as often as with estimated ones, so its abstention reflects small effective samples rather than estimation error. Its guarantees do not cover conditional shift, which leaves the covariates unchanged and goes undetected.

For practitioners, the results reduce to five habits:
\begin{itemize}
\item test UQ along the variable of deployment;
\item report sharpness, subgroup coverage and point error with coverage;
\item split calibration data by deployment unit, not by row;
\item keep tree-based intervals away from extrapolation;
\item treat abstention, and the effective sample size behind it, as part of the output.
\end{itemize}

No amount of source data or source validation can certify a model beyond its data; that knowledge has to come from physics, from new measurements, or from the decision not to predict.

\paragraph{Code and data availability} The benchmark code, the tuned and fixed-setting results, the ablations and a notebook that reproduces every table and figure will be made publicly available on publication. All seven datasets are public: Concrete, Energy, CCPP, Yacht, Naval and Air Quality from the UCI Machine Learning Repository, and C-MAPSS from the NASA Prognostics Center of Excellence Data Repository.


\printcredits

\bibliographystyle{cas-model2-names}

\bibliography{references}



\end{document}